\pdfoutput=1

\documentclass[11pt]{article}

\usepackage[final]{acl}

\usepackage{times}
\usepackage{latexsym}
\usepackage{booktabs}
\usepackage[T1]{fontenc}
\usepackage[utf8]{inputenc}
\usepackage{microtype}
\usepackage{inconsolata}
\usepackage{graphicx}
\usepackage{subcaption}

\usepackage{soul}
\usepackage{amssymb}
\usepackage{amsmath}
\usepackage{xcolor}
\usepackage{multicol}
\usepackage{booktabs}
\usepackage{moresize}
\usepackage{array}
\usepackage{arydshln}
\usepackage{multirow}
\usepackage{colortbl}
\usepackage{tocloft}
\usepackage{listings}
\usepackage{enumitem}
\usepackage{float}
\usepackage{bbm}
\usepackage{dsfont}
\usepackage{arydshln} 
\usepackage{tabularx}

\usepackage{placeins}

\newcommand{\TELLME}{\textsc{Tellme}}
\newcommand{\TELLMES}{\textsc{Tellme }}
\newcommand{\instpt}{\textsc{InstPT}}

\DeclareRobustCommand{\indicator}{\mathds{1}}
\newcommand{\CPTSFT}{\textsc{cpt+it}}
\newcommand{\CPT}{\textsc{cpt}}

\newcommand{\rev}[1]{\textcolor{black}{#1}}
\newcommand{\ihwon}[1]{\textcolor{black}{#1}}

\definecolor{insptColor}{RGB}{60, 215, 190}
\newcommand{\instptBG}[1]{%
  \begingroup
    \sethlcolor{insptColor}\hl{#1}%
  \endgroup
}

\definecolor{TELLMEColor}{RGB}{215, 190, 60}
\newcommand{\TELLMEBG}[1]{%
  \begingroup
    \sethlcolor{TELLMEColor}\hl{#1}%
  \endgroup
}

\definecolor{lightblue}{RGB}{230, 245, 255}
\newcommand{\blue}{\cellcolor{lightblue}}
\definecolor{lightgreen}{RGB}{230, 255, 245}
\newcommand{\green}{\cellcolor{lightgreen}}

\definecolor{mavblue}{RGB}{180, 195, 245}
\newcommand{\darkblue}{\cellcolor{mavblue}}
\definecolor{mavgreen}{RGB}{180, 245, 195}
\newcommand{\darkgreen}{\cellcolor{mavgreen}}

\definecolor{LightGray}{gray}{0.7}
\usepackage{kotex}
\usepackage{moresize}

\makeatletter
\def\adl@drawiv#1#2#3{%
        \hskip.1\tabcolsep
        \xleaders#3{#2.5\@tempdimb #1{1}#2.5\@tempdimb}%
                #2\z@ plus1fil minus1fil\relax
        \hskip.5\tabcolsep}
        
\newcommand{\cdashlinelr}[1]{%
  \noalign{\vskip\aboverulesep
           \global\let\@dashdrawstore\adl@draw
           \global\let\adl@draw\adl@drawiv}
  \cdashline{#1}
  \noalign{\global\let\adl@draw\@dashdrawstore
           \vskip\belowrulesep}}
\makeatother

\title{TELLME: Test-Enhanced Learning for Language Model Enrichment}

\newcommand\CoauthorMark{\footnotemark[\value{footnote}]}

\author{
    Minjun Kim$^1$\thanks{~~~These authors contributed equally to this work}\hspace{2.5mm}
    Inho Won$^1$\protect\CoauthorMark\hspace{2.5mm}
    Hyeonseok Lim$^1$\hspace{2.5mm}
    MinKyu Kim$^2$\hspace{2.5mm}\\
    \bf Junghun Yuk$^1$\hspace{2.5mm}
    Wooyoung Go$^3$\hspace{2.5mm}
    Jongyoul Park$^2$\hspace{2.5mm}
    Jungyeul Park$^1$\hspace{2.5mm}
    KyungTae Lim$^1$\thanks{~~~Corresponding Author} \\
    $^1$Korea Advanced Institute of Science and Technology \\
    $^2$Seoul National University of Science and Technology \\
    $^3$National Security Research Institute\\
    \texttt{\{mjkmain, inho.won, ktlim\}@kaist.ac.kr} \\
}

\begin{document}
\maketitle
\begin{abstract}

Continual pre-training (CPT) has been widely adopted as a method for domain \rev{adaptation} in large language models. However, CPT has consistently been accompanied by challenges, such as the difficulty of acquiring large-scale domain-specific datasets and high computational costs. In this study, we propose a novel method called Test-Enhanced Learning for Language Model Enrichment (\TELLME) to alleviate these issues. \TELLME\ leverages the Test-Enhanced Learning (TEL) principle, whereby the model’s \rev{training} efficiency is improved using quizzes during training. It integrates this principle with CPT, thereby promoting efficient domain-specific knowledge acquisition and long-term memory retention. Experimental results demonstrate that \TELLME\ outperforms existing methods by up to 23.6\% in the financial domain and achieves a 9.8\% improvement in long-term memory retention. The model and \rev{\TELLME\ dataset} are available at \url{huggingface.co/anonymous4459}.

\end{abstract}

\addtocontents{toc}{\protect\setcounter{tocdepth}{0}} 
\section{Introduction}\label{sec1:Introduction}

Recently released Large Language Models (LLMs) have demonstrated exceptional performance across various Natural Language Processing (NLP) tasks and are widely utilized~\citep{achiam2023gpt,brown2020language}. However, to tailor these models to specific domains or \rev{task-specific demands}, it is necessary to incorporate domain-specific knowledge through continual learning (CL). Depending on the objective, CL approaches have been proposed utilizing continual pre-training (CPT), instruction tuning (IT), or reinforcement learning (RL)~\citep{gururangan2020don,ouyang2022training,taori2023stanford}. Among these approaches, additional training using CPT has been recognized as an effective method for developing domain-specific LLMs by incorporating the intrinsic knowledge of the target domain. Nevertheless, CPT presents several challenges: (1) acquiring a large volume of domain-specific training data is often difficult, and (2) the training process requires substantial computational resources~\citep{wu2024continual}.

\begin{figure}[t]
  \centering
  \includegraphics[width=\linewidth]{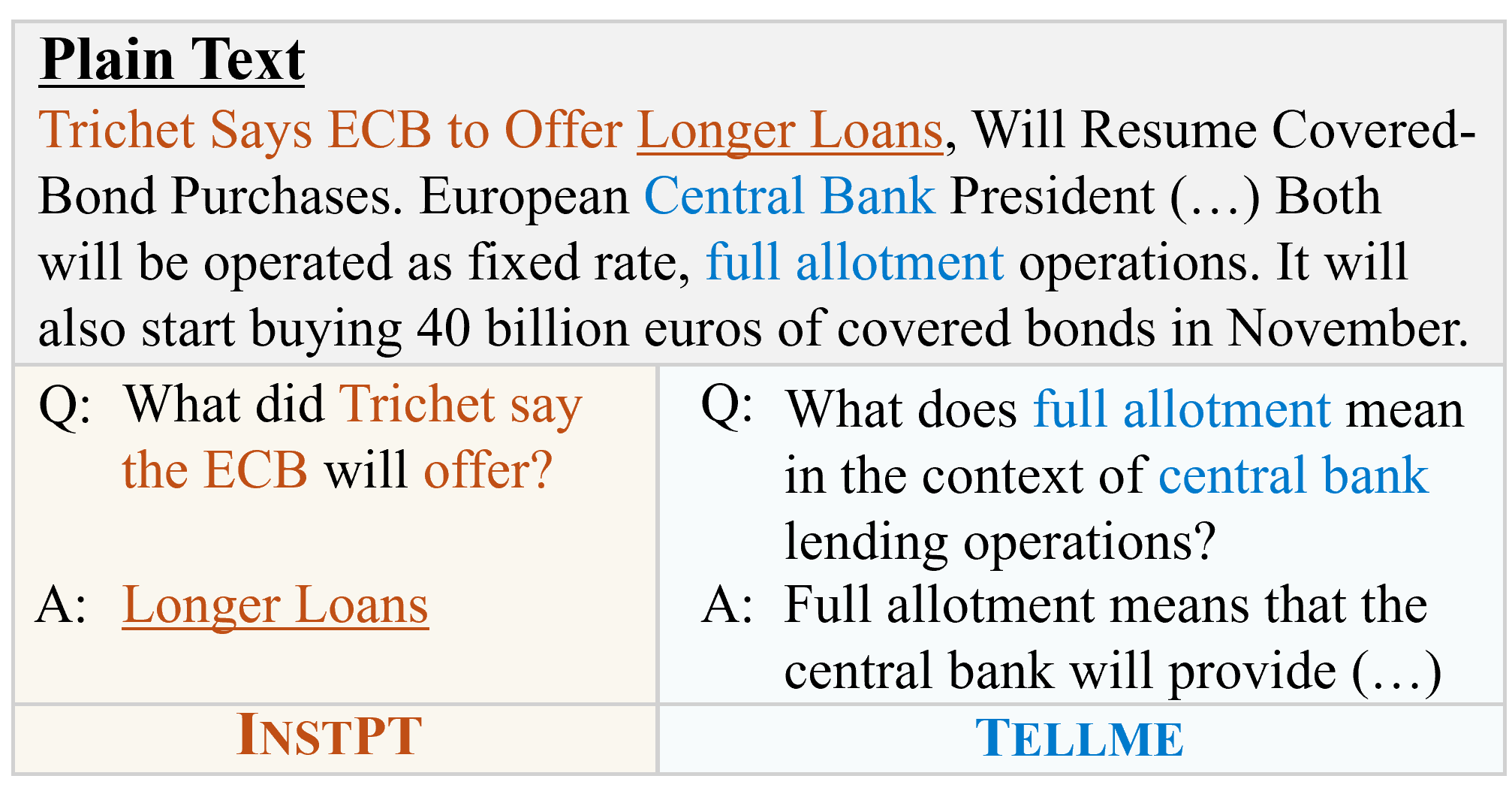}
    \caption{\rev{Examples of QA pairs produced with the \instpt\ and \TELLME\ methods. Whereas \instpt\ adopts a reading-comprehension style QA that extracts answers directly from the context, \TELLME\ reinforces knowledge through in-depth QA.
    }}
    \label{fig:inst_tel_qa}
\end{figure}

To address these issues, various methods have been proposed to perform effective CPT by further processing or augmenting CPT data. A noteworthy advancement is the shift away from the conventional approach, where IT is conducted sequentially after CPT, toward methods that incorporate IT data directly during the CPT process. This integrated approach has demonstrated promising results~\citep{cheng-etal-2024-instruction,jiang2024instruction,ke2025demystifying}. A representative example is \instpt, as illustrated in Figure~\ref{fig:inst_tel_qa}, where CPT is conducted simultaneously with QA samples related to plain text~\citep{cheng-etal-2024-instruction}. This approach effectively guides the models in encoding plain text knowledge more efficiently.

The previously proposed methods share a common feature: they integrate testing into the training process, resembling the Test-Enhanced Learning (TEL) framework in educational psychology~\citep{roediger2006test}. TEL has been shown to improve long-term retention by incorporating testing during the learning process. However, existing approaches such as \instpt\ and pre-instruction tuning~\citep{jiang2024instruction} deviate from the effective testing strategies suggested by TEL. Research shows that open-ended explanatory responses, rather than simple recall or multiple-choice formats, yield stronger long-term retention~\citep{larsen2008test,francis2020classroom}. In contrast, the question–answering format in \instpt, illustrated in Figure~\ref{fig:inst_tel_qa}, is largely constrained by the given text, limiting its ability to elicit internal knowledge.

Motivated by these findings, we hypothesize that adapting TEL’s principle of intrinsic-knowledge recall to CPT can improve both the efficiency of knowledge acquisition and the durability of learned representations. To test this hypothesis, we propose Test-Enhanced Learning for Language Model Enrichment (\TELLME). As illustrated in Figure~\ref{fig:inst_tel_qa}, \TELLME\ extends conventional CPT by jointly training plain text with descriptive QA samples that require explanatory reasoning beyond the given context. We construct 100K domain-specific \TELLME\ samples using GPT-4o-mini in a cost-efficient manner, ensuring high question diversity to stimulate the model’s intrinsic knowledge.

We evaluate \TELLME\ through domain-specific continual training and long-term retention experiments. Training datasets were built for the financial and medical domains, and additional training was performed using models of various scales, including LLaMA~\citep{dubey2024llama} and SmolLM~\citep{allal2025smollm2}. Experimental results show that \TELLME\ yields up to a 23.6\% improvement in financial comprehension benchmarks over CPT+IT baselines and achieves a 9.8\% gain in long-term retention compared with standard CPT. Our main contributions are summarized as follows:
\begin{itemize}[noitemsep]
\vspace{-0.2cm}
\item We introduce \TELLME, a continual pre-training framework that enhances knowledge acquisition and long-term retention in LLMs.
\item We present a cost-efficient pipeline for generating large-scale, diverse QA data for domain-specific continual training.
\item We empirically validate \TELLME\ on financial and medical domains, demonstrating significant gains over conventional CPT and CPT+IT methods.
\end{itemize}

\section{Related Work}\label{sec2:Related-work}

In this section, we introduce the foundational concepts underlying the proposed \TELLME\ method: (1) test-enhanced learning, (2) continual learning in LLMs, and (3) QA-based continual learning.

\subsection{Test-Enhanced \texttt{Learning for Human}} 


Test-Enhanced Learning(TEL)~\citep{roediger2006test}, one of the domain-optimized learning methods used by humans, is a concept studied in cognitive psychology. Unlike the general perception that tests merely serve as assessment tools, TEL has been shown to actively facilitate learning and enhance memory retention. This phenomenon is known as the testing effect, and research has demonstrated that it exhibits synergistic benefits, particularly when combined with concept mapping, which involves describing the relationships between distinct pieces of knowledge~\citep{francis2020classroom}. Additionally, studies have shown that TEL contributes to the long-term retention of domain-specific information.

Due to these advantages, TEL has been applied across various domains~\citep{butler2007testing, brame2015test}. Numerous studies in medical education have reported that exams requiring short-answer or descriptive responses rather than multiple-choice questions are more effective in reinforcing learning~\citep{larsen2013comparative, zheng2022impact, raksakietisak2024test}.

\subsection{Continual \texttt{Learning for LLMs}}

Domain optimization methods for LLMs primarily leverage continual learning, which enables them to adapt to new data distributions or domains. Within this framework, various approaches have been explored to enhance specific domains~\citep{xie-etal-2024-efficient, ke2025demystifying}, tasks~\citep{gururangan-etal-2020-dont}, and languages~\citep{fujii2024continual}, as well as to keep models updated with newly emerging information~\citep{lazaridou2021mind, su-etal-2023-efficient}. Specific examples of domain expansion can be found in the Appendix~\ref{Appendix:case studies}.

\subsection{QA-based Continual \texttt{Learning for LLMs}}
The concepts of Test-Enhanced Learning (TEL) in humans and additional training methods for LLMs have recently converged in QA-based CPT. A notable example is the pre-instruction tuning method proposed by ~\citet{jiang2024instruction}, which integrates plain text and QA samples into a mixed training process, enabling the model to learn both passages and QA pairs simultaneously. This approach has been reported to facilitate the efficient internalization of knowledge from plain text during training.

Another notable study has been proposed from the perspective of knowledge retention. \citet{ke2025demystifying} observed that performance degradation occurs due to the loss of instruction-following ability during continual pre-training and proposed a method that utilizes a mixture of the pre-training corpus and the instruction-following dataset to address this issue. Meanwhile, research has also been conducted on enhancing specific languages through QA-based CPT. For example, \citet{chen2024towards} proposed a CPT method targeting English and Chinese, leveraging synthetic QA data to improve model performance in the scientific domain.

Furthermore, QA-based CPT has been explored to strengthen reading comprehension abilities. \citet{cheng2023adapting, cheng-etal-2024-instruction} introduced \instpt, an instruction pre-training approach that utilizes template-based synthetic QA data to enhance specific tasks. This method has demonstrated notable improvements in the medical domain.
\section{\TELLME}\label{sec4:Proposed-method}
Test-Enhanced Learning for Language Model Enrichment (\TELLME) is a method designed to enhance the efficiency of knowledge acquisition and ensure long-term retention of learned knowledge by utilizing QA data during CPT. To implement this, this study describes the \TELLME\ method through \rev{(1) recap of language modeling,} (2) \rev{question-and-answer} generation from plain text, and (3) the design of a \rev{training} framework.

\subsection{\rev{Language Modeling}}\label{sec3:preliminary}
To facilitate \rev{better} understanding of the proposed \TELLME, \rev{we summarize the key concepts of causal language modeling (CLM)}, pre-training (PT), and instruction tuning (IT), which constitute the fundamental training methods of CPT.


\paragraph{Causal Language Modeling (CLM)}
LLMs are optimized using the CLM objective, which predicts the next token based on the preceding context. This objective can be formulated as follows:
\begin{equation}
    \mathcal{L}_{\textsc{clm}}(\theta) = -\frac{1}{K}\sum_{i=1}^N \indicator(x_i)\log P(x_{i}|x_{<i};\theta)\\
\label{eq:causal_loss}
\end{equation}

Here, $\theta$ denotes the model parameters, $N$ is the sequence length, $x_i$ is the $i$-th token, and $x_{<i}$ denotes all tokens preceding $x_i$. 
The normalization term is given by $K = \sum_{i=1}^N \indicator(x_i)$, which accounts for the total number of tokens contributing to the loss. The indicator function $\indicator(x_i)$ is defined as:
\begin{equation}
    \indicator(x_i) = \begin{cases}
            1 & \text{if $i$-th token included in loss}\\
            0 & \text{otherwise}
        \end{cases}
\end{equation}

Depending on the training dataset composition and indicator function configuration, this loss function can be categorized into PT and IT. 

\paragraph{Pre-Training (PT)} The dataset used for PT consists of large-scale textual corpora encompassing extensive general knowledge. These datasets typically comprise plain text at the sentence or document level. During PT, given an input token sequence $\mathbf{x} = (x_1, \ldots, x_N)$, the indicator function is set as $\indicator(x_i \in \mathbf{x}) = 1$ for all tokens, ensuring that every token contributes to the loss computation. This enables the optimization of a probabilistic model $P(x_i | x_{<i}; \theta)$ over the entire corpus.

\paragraph{Instruction Tuning (IT)} In contrast, IT employs a relatively small, structured dataset that prioritizes learning task-specific response patterns (e.g., translation, summarization) rather than acquiring broad knowledge. In this case, training sample $\mathbf{x}$ consists of an input prompt $\mathbf{p}$ concatenated with an output $\mathbf{o}$: represented as $\mathbf{x} = (\mathbf{p}, \mathbf{o})$. During IT, only tokens belonging to $\mathbf{o}$ contribute to the loss computation. This is implemented by setting the indicator function such that $\indicator(x_i \in \mathbf{p}) = 0$ and $\indicator(x_i \in \mathbf{o}) = 1$, ensuring that the model learns to generate appropriate responses while disregarding loss contributions from the input prompt.


\begin{table}[t]
\centering
\renewcommand{\arraystretch}{1}
\resizebox{.49\textwidth}{!}{
{
\begin{tabular}{@{}l@{}}
\hline
\textbf{Prompt} \\
\hline
\smash{\textbf{[\texttt{instruction}]}: Generate a Q\&A based on the following requirements} \\
1. Avoid \rev{direct question about given excerpt.}\\
2. \rev{Create question based on general domain knowledge.}\\
3. \rev{Ensure question can be answered independently of the excerpt.}\\

\textbf{[\texttt{input}]:} \\
The French banking bill prohibits high-frequency trading in (...)
\\
\hline
\end{tabular}}
}
\vspace{-2mm} 
\caption{A simplified prompt example for constructing the \TELLME\ dataset.}
\label{tab:instruction-data-dp}
\end{table}

\subsection{Dataset Curation for \TELLME}\label{sec4.1:data-curation}
As previously described, TEL has been reported to be particularly effective when (1) the questions are descriptive and (2) the answers require respondents to incorporate their own opinions (internal knowledge) along with factual information. Therefore, it is preferable to design questions that allow for diverse and unconstrained expression of opinions.
Accordingly, we first avoided reading comprehension-style questions that can be answered merely by referring to the plain text. Instead, we focused on generating QA pairs that, while related to the plain text, address new knowledge that cannot be directly found within the text. 
Table~\ref{tab:instruction-data-dp} provides a simple example of a QA generation prompt constructed based on these criteria. 
In the [\texttt{input}] of Table~\ref{tab:instruction-data-dp}, it is evident that the plain text pertains to a banking bill passed in France. Based on this, and following the rules proposed in the [\texttt{instruction}], a question such as
``How do high-frequency trading strategies impact market volatility?''
can be generated. This question establishes a conceptual connection (concept mapping) between ``high-frequency trading'' and ``market volatility'', two related pieces of knowledge that are not explicitly mentioned in the plain text. We hypothesize that this structure enhance long-term memory retention of knowledge in the respective domain.

In this study, we constructed the \TELLME\ dataset using the GPT4o-mini based on the proposed prompt. Generating 100K samples with GPT4o-mini cost approximately \$12 in total. The training data \rev{spans medical and financial domains, with each sample containing plain text and $M$ associated question-answer pairs.}
A concrete example of the \TELLME\ dataset is shown in Figure~\ref{fig:inst_tel_qa}, with detailed data samples and generation prompts provided in Appendices~\ref{Appendix:Tellme-Dataset-Examples-and-Clarification} and \ref{Appendix:TEL-Dataset-Info}. Finally, the generated data achieved an average score of 4.03 out of 5 in an LLM-as-a-judge quality evaluation based on relevance, clarity, and completeness. The detailed evaluation prompts and results for data quality assessment are provided in Appendix~\ref{app:dataset-quality}.

\subsection{Adapting TEL to Continual Learning}
The previously constructed \TELLME\ dataset samples follow the structure $\rev{\mathbf{X}} = (\mathbf{t}, \mathbf{q}, \mathbf{a})$, where $\mathbf{t}$ represents the token sequence of the plain text, and $\mathbf{q}$ and $\mathbf{a}$ correspond to the token sequences of the questions and their respective answers. For simplicity, this structure assumes $M=1$. When $M>1$, the structure can be extended through QA concatenation as $\rev{\mathbf{X}} = (\mathbf{t}, \mathbf{q}_1, \mathbf{a}_1, \ldots, \mathbf{q}_M, \mathbf{a}_M)$.

To explicitly reflect the testing effect, we utilize the \TELLME\ dataset, which includes both plain text and QA within a single sample. Specifically, in Equation~\ref{eq:causal_loss}, we configure the indicator function as $\indicator(x_i \in \mathbf{t} \cup \mathbf{a}) = 1$, $\indicator(x_i \in \mathbf{q}) = 0$. This ensures that the model is trained to predict only the plain text and answer components while excluding the question component $\mathbf{q}$ from the loss computation.

From the perspective of mixed training using both plain text and QA samples, this method serves as a natural extension of the conventional CLM approach, considering the CPT and IT training paradigms. Consequently, it incorporates the \TELLME\ framework. 


\begin{table*}[!h]
\centering
\small
\begin{tabular}{lccccccccc}
    \toprule
    \multirow{2}{*}{\textbf{Model}} & \multicolumn{4}{c}{\textbf{Finance}} & \multicolumn{4}{c}{\textbf{Medicine}}  &\multirow{2}{*}{\textbf{Average}}\\
    \cmidrule(r){2-5}\cmidrule(l){6-9}
                             & FOMC            & NIFTY           & MMLU-F            &         AVG.           &      HeadQA     &     MedMCQA     &      MMLU-C     &        AVG.            \\ 
    \midrule
               Llama-3.2-1B    &      22.04      &      30.38      &      39.76      &      \blue{30.73}      &      32.39      &      28.19      &      35.92      &      \blue{32.16}        & \green{31.45} \\
               + \CPT          &      22.04      &      25.09      &      40.01      &      \blue{29.05}      &      33.58      &      27.71      &      35.48      &      \blue{32.26}        & \green{30.66} \\               
               + \CPTSFT       &      24.49      &      27.74      &      38.84      &      \blue{30.36}      &      29.25      &      26.75      &      33.27      &      \blue{29.76}        & \green{30.06} \\
               + \instpt       &      28.17      &      27.34      &      37.92      &      \blue{31.15}      &      29.80      &      27.06      &      33.39      &      \blue{30.08}        & \green{30.61} \\
               \rowcolor{gray!30} 
               + \TELLME       &      29.74      &      30.69      &      39.96      &  \darkblue{\textbf{33.46}} &      33.55      &      28.21      &      36.11      &  \darkblue{\textbf{32.62}}   & \darkgreen{\textbf{33.04}} \\
    \midrule
               Llama-3.2-3B    &      22.04      &      29.41      &      47.47      &      \blue{32.97}      &      37.93      &      31.80      &      43.15      &      \blue{37.63}        & \green{35.30}\\
               + \CPT          &      22.04      &      20.75      &      47.15      &      \blue{29.98}      &      38.88      &      31.05      &      41.73      &      \blue{37.22}        & \green{33.60}\\               
               
               + \CPTSFT       &      21.79      &      25.47      &      45.36      &      \blue{30.87}      &      33.55      &      31.17      &      42.21      &      \blue{35.64}        & \green{33.26}\\
               + \instpt       &      28.79      &      19.60      &      44.68      &      \blue{31.03}      &      34.14      &      31.44      &      39.89      &      \blue{35.16}        & \green{33.10}\\
               \rowcolor{gray!30} 
               + \TELLME       &      26.02      &      27.29      &      48.31      &  \darkblue{\textbf{33.87}} &      38.99      &      32.01      &      43.67      &  \darkblue{\textbf{38.22}}   & \darkgreen{\textbf{36.05}}\\
    \midrule
               Llama-3.1-8B    &      22.04      &      23.72      &      53.45      &      \blue{33.07}      &      42.71      &      37.51      &      51.74      &  \blue{\textbf{43.99}}   & \green{38.53}\\
               + \CPT          &      29.53      &      30.16      &      52.65      &      \blue{37.44}      &      42.85      &      35.14      &      49.32      &      \blue{42.44}        & \green{39.94}\\               
               + \CPTSFT       &      23.51      &      22.40      &      48.33      &      \blue{31.41}      &      34.06      &      31.99      &      43.82      &      \blue{36.62}        & \green{34.02}\\
               + \instpt       &      34.40      &      27.30      &      49.25      &      \blue{36.99}      &      36.98      &      33.71      &      45.29      &      \blue{38.66}        & \green{37.83}\\
               \rowcolor{gray!30} 
               + \TELLME       &      31.38      &      31.40      &      53.69      &  \darkblue{\textbf{38.82}} &      43.00      &      35.60      &      49.29      &      \darkblue{42.63}        & \darkgreen{\textbf{40.73}}\\
    \midrule
               SmolLM2-1.7B    &      26.31      &      23.89      &      47.38      &      \blue{32.53}      &      36.83      &      29.76      &      39.89      &      \blue{35.49}        & \green{34.01}\\
               + \CPT          &      28.56      &      27.79      &      46.49      &      \blue{34.28}      &      36.61      &      29.64      &      39.59      &      \blue{35.28}        & \green{34.78}\\               
               
               + \CPTSFT       &      26.36      &      30.33      &      47.16      &      \blue{34.62}      &      36.47      &      29.74      &      40.92      &      \blue{35.71}        & \green{35.17}\\
               + \instpt       &      28.75      &      33.38      &      45.73      &  \blue{\textbf{35.95}} &      36.69      &      29.76      &      39.71      &      \blue{35.39}        & \green{35.67}\\
               \rowcolor{gray!30} 
               + \TELLME       &      29.37      &      30.61      &      47.46      &    \darkblue{35.82}        &      37.13      &      30.00      &      41.28      &  \darkblue{\textbf{36.14}}   & \darkgreen{\textbf{35.98}}\\
\bottomrule
\end{tabular}
\caption{Comparison of performance in the finance and medicine domains under different training methods.}
\label{tab:main_table}
\end{table*}

\section{Experiment}\label{sec5:Expreiment}
In this section, we present the quantitative evaluation procedure and criteria for the proposed \TELLME\ method, and analyze the experimental results based on the following research questions: (1) Does \TELLME\ method acquire domain knowledge more efficiently than existing approaches? and (2) Is it effective for long-term memory retention?

\subsection{Experimental settings}\label{subsec5.1:Experiment-environments}
In this study, we focused on the financial and medical domains, constructing datasets based on the method proposed in Section~\ref{sec4.1:data-curation} using 100k PubMed abstracts and 100k Bloomberg financial news articles. The experiments were conducted with the number of QA pairs per data sample to $M=3$. Both the medical and financial domains require specialized knowledge and have been primarily used in previous studies for performance validation based on CPT~\citep{pezeshkpour2025learning, phasook2024thaibkd}. 
The evaluation benchmarks for finance include FOMC~\citep{shah-etal-2023-trillion}, NIFTY~\citep{saqur2024nifty}, and MMLU-F(inance). For the medical domain, evaluations were conducted using HeadQA~\citep{vilares-gomez-rodriguez-2019-head}, MedMCQA~\citep{pmlr-v174-pal22a}, and MMLU-C(linic)~\citep{singhal2025toward}. All evaluations were conducted using the lm-evaluation-harness~\citep{eval-harness} for reproducibility. Appendix~\ref{Appendix:Training-Details-and-Hyperparameters} provides a detailed description of the benchmark datasets used for evaluation.

\begin{table*}[!h]
\centering
\small
\begin{tabular}{lccccccccc}
    \toprule
    \multirow{2}{*}{\textbf{Model}} & \multicolumn{4}{c}{\textbf{Finance}} & \multicolumn{4}{c}{\textbf{Medicine}}  &\multirow{2}{*}{\textbf{Average}}\\
    \cmidrule(r){2-5}\cmidrule(l){6-9}
                             & FOMC            & NIFTY           & MMLU-F            &         AVG.           &      HeadQA     &     MedMCQA     &      MMLU-C     &        AVG.            \\ 
    \midrule
               Llama-3.2-3B    &      22.04      &      29.41      &      47.47      &      \blue{32.97}      &      37.93      &      31.80      &      43.15      &      \blue{37.63}        & \green{35.30}\\
               + \CPT          &      22.04      &      20.75      &      47.15      &      \blue{29.98}      &      38.88      &      31.05      &      41.73      &      \blue{37.22}        & \green{33.60}\\               
               
               + \CPTSFT       &      21.79      &      25.47      &      45.36      &      \blue{30.87}      &      33.55      &      31.17      &      42.21      &      \blue{35.64}        & \green{33.26}\\
               + \instpt       &      28.79      &      19.60      &      44.68      &      \blue{31.03}      &      34.14      &      31.44      &      39.89      &      \blue{35.16}        & \green{33.10}\\
               \rowcolor{gray!30} 
               + \TELLME       &      26.02      &      27.29      &      48.31      &  \darkblue{\textbf{33.87}} &      38.99      &      32.01      &      43.67      &  \darkblue{\textbf{38.22}}   & \darkgreen{\textbf{36.05}}\\
\bottomrule
\end{tabular}
\caption{Comparison of performance in the finance and medicine domains under different training methods.}
\label{tab:main_table}
\end{table*}

The evaluations were conducted using state-of-the-art open-source LLMs with varying capabilities as the base models. Specifically, experiments were performed using Llama-\rev{\{}3.2-1B, 3.2-3B, 3.1-8B\rev{\}}~\citep{llama3modelcard} and SmolLM2-1.7B~\citep{allal2025smollm2}. \rev{To assess the effectiveness of the proposed method}, we compared the performance of the \TELLME\ method with existing approaches based on baseline models and four variations of the training methods:
\begin{itemize} [noitemsep]
\vspace{-0.2cm}
    \item + \textsc{cpt}: Refers to the model that has undergone continual pre-training on domain-specific texts.
    \item + \CPTSFT: The model instruction-tuned on a QA dataset based on the + \textsc{cpt} model ~\citep{yang2024pllamaopensourcelargelanguage, chen2023meditron70bscalingmedicalpretraining, colombo2024saullm7bpioneeringlargelanguage}.
    \item + \instpt: This model is trained based on the template-based QA generation approach proposed by \citet{cheng-etal-2024-instruction}. Specifically, an average of 5.8 short-form QA pairs is generated for the finance domain, while an average of 1.25 long-form QA pairs is generated for the medical domain. During the subsequent CPT, the plain text and QA datasets are concatenated, and the loss is computed over all the tokens.  In this case, the indicator function in Equation~\ref{eq:causal_loss}, $\indicator(x_i \in \mathbf{X}) = 1$. Further implementation details regarding \instpt\ can be found in Appendix~\ref{Appendix:Detailed-InstPT}.
    \item + \TELLME: The model trained using the proposed \TELLME\ method. 
    In this approach, each sample includes both plain text and QA pairs; however, 
    only the plain text and answer tokens are used when computing the loss.
\end{itemize}
Here, the CPT and IT stages of the \CPTSFT\ model utilized the plain text and QA samples from the \TELLME\ dataset, respectively. \rev{Consequently, both the \TELLME\ and \CPTSFT\ models see the same total number of tokens from the plain-text and QA data. However, \CPTSFT\ performs two seperated forward-backward passes, whereas \TELLME\ processed each mixed sample in a single pass.} Detailed information on the models and the training hyperparameters can be found in Appendix~\ref{Appendix:Training-Details-and-Hyperparameters}. 

\subsection{Experiment Results}\label{subsec4.2:Experiment-Results}

\paragraph{Overall}  Table~\ref{tab:main_table} presents the performance of the baseline, \textsc{cpt}, \CPTSFT, \instpt, and \TELLME\ models in the financial and medical domains. Overall, the \TELLME\ approach achieves the highest average performance. Notably, it outperformed the commonly used \CPTSFT\ method by 10.0\%, demonstrating a significant improvement. Given that both \TELLME\ and \CPTSFT\ are trained on the same number of tokens, this result suggests that \TELLME\ enables more efficient learning of domain-specific knowledge compared to existing methods. Furthermore, \TELLME\ surpasses \instpt\ by 6.3\% overall across both the financial and medical domains, indicating that the incorporation of open-ended, free-form QA has a positive impact.

\paragraph{\TELLME\ for Domain Adaptation}
How does \TELLME\ perform across distinct domains? The experimental results show that the \TELLME\ method achieves strong performance in both the finance and medical domains, with particularly notable improvements in the finance domain.
In the finance domain, \TELLME\ consistently outperformed the baseline model across all tested models, achieving an average performance gain of approximately 9.8\%. Moreover, despite the \instpt\ method training on nearly twice as much QA data as \TELLME, the \TELLME\ approach still achieved higher scores in all models, except for SmolLM2-1.7B. Overall, \TELLME\ outperformed \instpt\ by an average of 5.1\% across all models.
In the medical domain, \TELLME\ demonstrated an average improvement of 0.09 points over the baseline model and outperformed \instpt\ by an average of 2.58 points.

\begin{figure}[!t]
  \includegraphics[width=\linewidth]{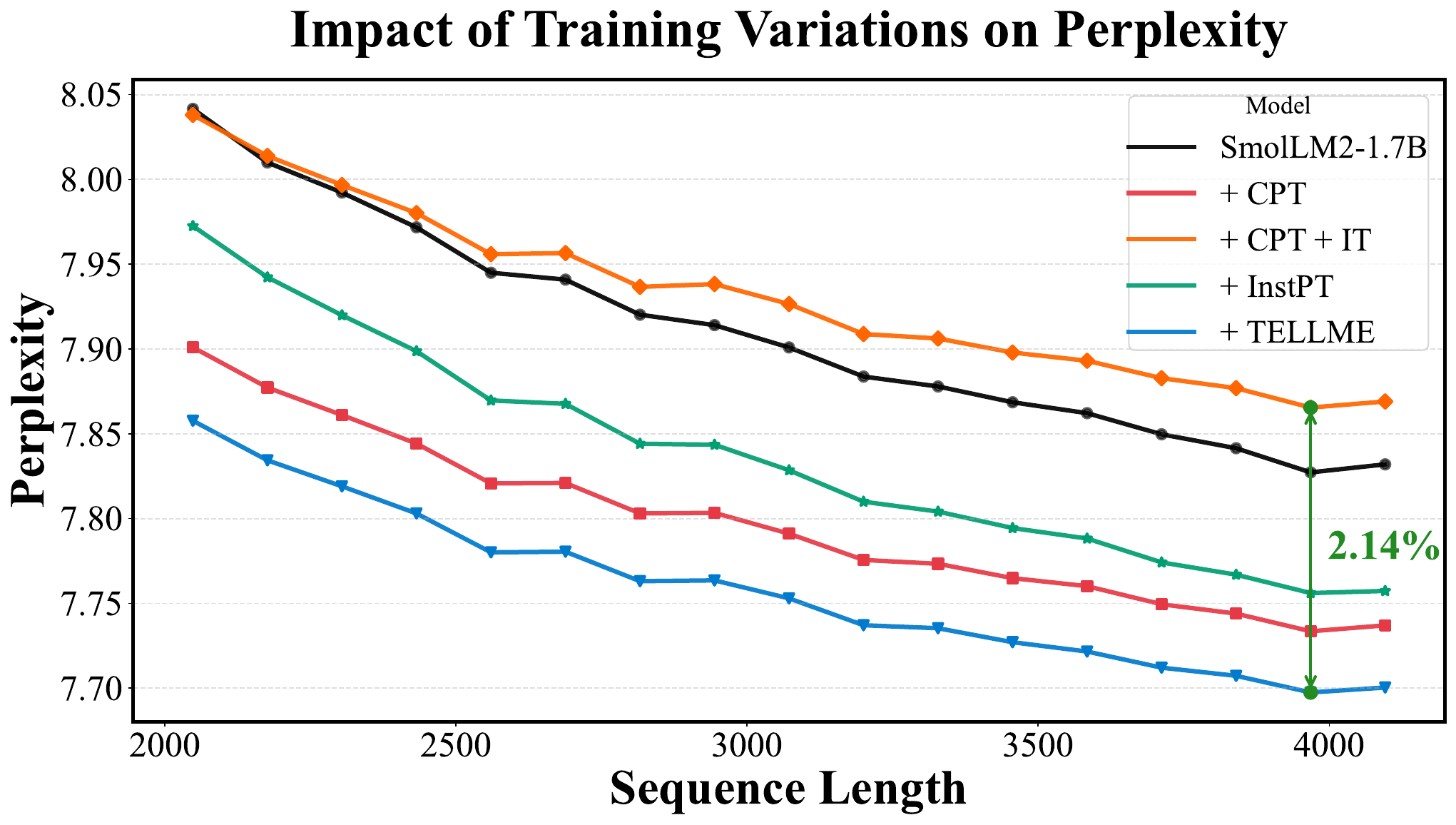}
    \caption{Perplexity comparison on the medical dataset for different training approaches. The baseline model shows the highest perplexity, while the proposed \TELLMES achieves the lowest perplexity across all lengths, indicating improved model performance.}
  \label{fig:perplexity-medical-smollm2-1.7B}
\end{figure}

\paragraph{Perplexity in Domain Adaptation} 
Figure~\ref{fig:perplexity-medical-smollm2-1.7B} presents a comparison of perplexity(PPL) for five training strategies, including the proposed \rev{\TELLME} method, in the medical domain. For both training and evaluation, plain text from the PubMed dataset was employed, with 100k and 4k disjoint data samples, respectively, to ensure a fair evaluation. The results indicated that the pre-trained baseline model (SmolLM2-1.7B), which was not subjected to domain adaptation, exhibited relatively high PPL across all sequence lengths. In contrast, the model trained solely on plain text (+\textsc{cpt}) tended to demonstrate lower PPL, suggesting a positive effect on domain adaptation. However, the model that underwent additional IT with QA data after plain text training (+\CPTSFT) unexpectedly exhibited the highest PPL. This observation is interpreted as the QA-focused fine-tuning phase conducted at the end, diluting the plain text representational capacity. 
On the other hand, the +\TELLME\ model, which combines plain text and QA within a single data sample achieved the lowest PPL across all sequence length intervals. This suggests that, compared to the +\textsc{cpt} model, the additional QA component in the +\TELLME\ model exerts a positive impact on domain adaptation, and that TEL efficiently acquires domain-relevant representations.

\subsection{The Impact of TEL on Long-Term Retention}

\begin{figure}[t]
  \includegraphics[width=\linewidth]{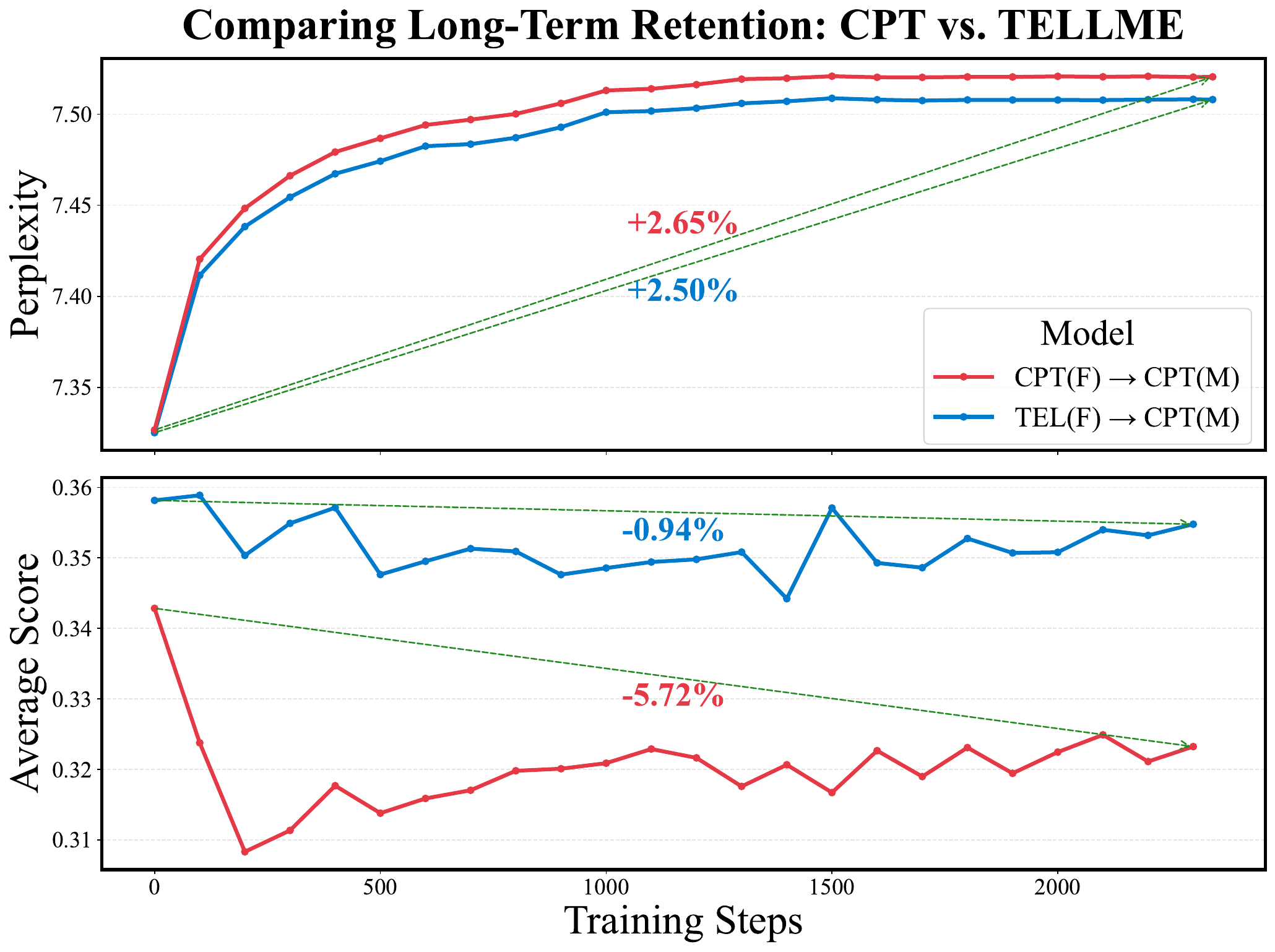}
    \caption{Performance of the finance domain after overwriting with medicine data.}
  \label{fig:finance_retention}
\end{figure}

We conducted two experiments using SmolLM2-1.7B to assess the effectiveness of the TEL technique in retaining long-term \rev{domain} knowledge. Figure~\ref{fig:finance_retention} displays a comparison between two models (CPT(F)$\to$CPT(M) and TEL(F)$\to$CPT(M)) in
terms of training-step PPL \rev{measured on the Bloomberg corpus} (top) and performance based on a financial benchmark (bottom). Our experimental setup compared models trained in the following two phases: 
\begin{itemize}[noitemsep]
\vspace{-0.2cm}
    \item [1.] First, we trained two initial models using finance domain data: one with \CPT\ and the other with \TELLME\ (CPT(F) and TEL(F)). 
    \item [2.] Subsequently, we further trained both models on medical domain data with \CPT\ for 3 epochs, yielding the final models: CPT(F)$\to$CPT(M) and TEL(F)$\to$CPT(M). 
\end{itemize}
\vspace{-0.2cm}

In this experiment, we analyzed the long-term retention of previously \rev{acquired} knowledge (Finance) by comparing the evaluation results on the target domain (Finance) between the CPT(F)$\to$CPT(M) model and the TEL(F)$\to$CPT(M) model.

\paragraph{Perplexity Results on Finance}

The top graph in Figure~\ref{fig:finance_retention} compares the PPL of the two models. The evaluation was conducted on 4K samples from the Bloomberg corpus, each with a sequence length of 4K, using a portion of the corpus that was not included in the training data. The experimental results show that the TEL-based model achieved a lower PPL than the CPT-based model, indicating superior performance. Specifically, in terms of the PPL increase rate, the TEL-based model exhibited a \rev{2.5}\% increase, whereas the CPT-based model showed a \rev{2.65}\% increase. Given that lower PPL indicates greater confidence in predicting the next token, these results suggest that the TEL-based model maintains a higher probability of generating content from its previously trained domain, despite additional training in an unrelated domain. This implies that compared to conventional \rev{CPT} methods, the TEL approach better preserves prior domain knowledge even after cross-domain adaptation.

\paragraph{\rev{Benchmark} Results on Finance}
The bottom graph in Figure~\ref{fig:finance_retention} compares the average performance of the TEL(F)$\to$CPT(M) and CPT(F)$\to$CPT(M) models on the finance benchmark. Observing the range between 0 and 250 training steps, the CPT(F)$\to$CPT(M) model exhibits a notable decline in finance domain comprehension early in the training process on the medical dataset. In contrast, the TEL(F)$\to$CPT(M) model shows more stable retention of finance domain knowledge, even after additional training on medical data. As a result, while the CPT-based approach suffered a \rev{5.72}\% decline in performance relative to its initial state, the TEL-based approach showed only a \rev{0.94}\% reduction, indicating superior knowledge retention. Additionally, despite starting with an initial performance 1.53 points higher than that of the CPT-based model, the TEL-based model maintained its knowledge more effectively throughout the training, ultimately achieving a 9.8\% higher final performance (equivalent to a 3.15-point increase) compared to the CPT-based approach. These findings suggest that, even when trained on an out-of-domain dataset, the TEL method preserves the knowledge of the target domain more effectively than conventional pre-training methods. This indicates that TEL has a positive impact on long-term retention, contributing to greater stability in learned domain knowledge.
\section{Ablation Study and Analysis}
\begin{table}[t]
    \resizebox{\columnwidth}{!}{
    \begin{tabular}{lccccc}
        \toprule
        \textbf{Model}       &      Inv-TEL      &        CPT        &       PIT         &       TEL-Q/L     &    \blue{\textbf{TEL}}         \\
        \midrule
        Llama-3.2-1B         &       31.20       &    29.05          &    31.81           &  \textbf{33.73}  &              \blue{33.46}    \\
        Llama-3.2-3B         &       31.95       &    29.98          &    33.69           &    33.06         &     \blue{\textbf{33.87}}    \\
        Llama-3.1-8B         &       33.83       &    37.44          &    37.61           &    38.69         &     \blue{\textbf{38.82}}    \\
        SmolLM2-1.7B         &       33.69       &    34.28          &    35.63           &    34.80         &     \blue{\textbf{35.82}}    \\
        \midrule
        Average              &      32.67        &    32.69          &    34.69           &    35.07          &    \blue{\textbf{35.49}}    \\   
        \bottomrule
    \end{tabular}
    }
    \caption{A performance comparison for various training methods utilizing QA data in the financial domain.}    \label{tab:anl-table}
\end{table}

This section provides an in-depth analysis of \TELLME, focusing on performance variations across different TEL application strategies. Table~\ref{tab:anl-table} presents a comparison of model performance across various training techniques in the finance domain. In this table, 
\rev{`TEL' refers to the proposed \TELLME\ method, and `Inv-TEL' denotes a training strategy in which the input data is structured as $\mathbf{X}=(\mathbf{q}, \mathbf{a}, \mathbf{t})$, where QA samples precede plain text.}
In this case, the loss function is computed in the same manner as in the \TELLME\ approach, where all tokens, except for the question, are treated as targets during CPT. Additionally, `PIT’ included as a comparative method, follows the approach proposed by~\citet{jiang2024instruction}. It consists of a two-stage training process: first, IT is conducted using a QA dataset, and then CPT is performed on the trained model by treating the QA dataset and plain text data as independent samples. During this process, the loss calculation for the questions is excluded. Lastly, `TEL-Q/L’ represents a variation of the \TELLME\ method in which the loss calculation is applied to all tokens, including questions, during CPT. See Appendix~\ref{appendix:d-4} for experimental details.

\paragraph{Performance on Test QA Placement}

\ihwon{Motivated by PIT's approach, we tested Inv-TEL to examine if positioning QA before plain text impacts performance.} As shown in Table~\ref{tab:anl-table}, the model trained using the Inv-TEL method exhibited a performance that was approximately 2.82 and 0.02 points lower than those of the TEL and CPT methods, respectively. These results suggest that even when TEL is applied, the positioning of the QA pair can significantly affect model performance, thereby demonstrating that the proposed \TELLME\ method effectively leverages this strategy.

\paragraph{Performance on QA Dataset Utilization Method}

The QA data proposed in this study can be utilized as independent samples in CPT alongside plain text. Alternatively, QA data can be incorporated within a single sample, along with plain text, for training purposes. How does performance differ when QA and plain text are treated as separate samples? As described earlier, the PIT in Table~\ref{tab:anl-table} represents a model trained with CPT by combining QA and plain text as independent samples. Compared to TEL, this model exhibited approximately 0.8 points lower performance. Ultimately, the results indicate that integrating QA and plain text within a single training sample yields higher efficiency than treating them as independent samples.

\paragraph{The Impact of Question Prediction Loss} 

Would excluding loss calculation for questions improve model performance? As shown in Table~\ref{tab:anl-table}, TEL-Q/L exhibited approximately 0.42 points lower performance compared to TEL. TEL-Q/L computes the loss for all tokens, including the questions, whereas TEL calculates the loss only for tokens excluding the questions. These results suggest that the proposed \TELLME\ training method efficiently learns and utilizes key information, ultimately leading to improved performance.

\paragraph{Efficiency of \TELLME}
\ihwon{The CPT has a limitation in which its training process requires substantial computational resources. To investigate whether \TELLME\ can alleviate this issue, we conducted an experiment measuring PPL with respect to training steps. The experimental results showed that \TELLME\ achieved the same PPL with a 1.4 times faster compared to CPT. 
 }\rev{Further details and results of the experiment are provided in the Appendix~\ref{further-anal}.}

\begin{table}[!h]
\centering
\small
\begin{tabular}{lcccc}
    \toprule
    \multirow{2}{*}{\textbf{Model}} & \multicolumn{4}{c}{\textbf{Finance}}\\
    \cmidrule(r){2-5}
                             & FOMC            & NIFTY           & MMLU            &         AVG.           \\ 
    \midrule
               Llama-3.2-3B    &      22.04      &      29.41      &      47.47      &      \blue{32.97}   \\
               
               + \instpt       &      28.79      &      19.60      &      44.68      &      \blue{31.03}   \\
               + \TELLME-(M)  &  24.01     &     30.06      &      45.84      &      \blue{33.30}   \\
               + \TELLME-(S) &      26.72      &    27.47      &      46.74      &      \blue{33.64}   \\
               \rowcolor{gray!30} 
               + \TELLME       &      26.02      &      27.29      &      48.31      &  \darkblue{\textbf{33.87}}\\

\bottomrule
\end{tabular}
\caption{Performance of \TELLME\ in financial domain across synthesizers. (M) and (S) indicate datasets generated by Mistral and self-generated datasets, respectively.}
\label{tab:synthesizer_analysis}
\end{table}

\paragraph{\rev{Impact of the Synthesizer in the \TELLME}}

\rev{The \textsc{Tellme} dataset was primarily constructed using the GPT4o-mini model. However, employing such a sophisticated model in dataset construction raises questions about whether performance improvements genuinely stem from the efficiency of the proposed \TELLME\ approach or merely reflect knowledge distillation from a superior model.} \rev{To investigate this perspective, we conducted additional experiments using an alternative synthesizer, specifically the Mistral-7B model utilized in the InstPT approach, to generate the \TELLME\ dataset. As shown in Table \ref{tab:synthesizer_analysis}, the Llama-3.2-3B model trained on the Mistral-generated dataset (\TELLME-(M)) achieved an average score of 33.30, outperforming the \instpt\ baseline by 2.27 points in the financial domain.} \rev{Furthermore, we explored the effectiveness of self-generated datasets, where the \TELLME\ dataset was generated by the target model itself (\TELLME-(S)). Results indicate that the self-generated dataset notably improved performance, 2.61 points higher than the \instpt\ baseline.} \rev{These results underscore that while employing a powerful synthesizer like GPT4o-mini yields superior performance, the \TELLME\ methodology remains robust and effective even when less advanced synthesizers or self-generated datasets are utilized.}

\paragraph{Multilingual Generalization to Korean}
\begin{table}[!h]
\centering
\scriptsize
\setlength{\tabcolsep}{3pt}
\begin{tabular}{@{}lcccccc@{}}
\toprule
Model & BoolQ & COPA & HLSW. & SENT. & WIC & Avg. \\
\midrule
Base              & 0.522 & 0.477 & 0.422 & 0.524 & 0.529 & 0.495 \\
\textsc{cpt}      & 0.571 & 0.632 & 0.534 & 0.536 & 0.515 & 0.546 \\
\textsc{tellme-ko} & 0.610 & 0.585 & 0.422 & 0.730 & 0.550 & \textbf{0.579} \\
\bottomrule
\end{tabular}
\caption{Benchmark results on KoBEST for the OLMo-1B model and its variants fine-tuned with \textsc{tellme-ko}.}
\label{tab:expr-tellme-short-ko}
\end{table}

We further investigated whether the proposed \TELLMES framework generalizes beyond English. To this end, we generated Korean \TELLMES data (\textsc{tellme-ko}) following the English setting. We evaluated Korean performance using the OLMo2-1B model on the KoBEST~\cite{jang2022kobest} benchmark, aiming to assess how effectively \TELLMES can enhance Korean proficiency in models that originally lack any Korean capability. As shown in Table~\ref{tab:expr-tellme-short-ko}, \textsc{tellme-ko} achieves a remarkable +8.4-point improvement in average accuracy, with over +20-point gains on the sentence understanding (SENT) task. These results highlight that the proposed \TELLMES framework can augment knowledge in a language-agnostic manner. Additional experiments across different Korean models, scales, and data generation methods are presented in Appendix~\ref{app:tellme-ko}.

\section{Conclusion}
In this study, we propose the \TELLME\ (Test-Enhanced Learning for Language Model Enrichment) technique, which offers an effective method for continual pre-training of large language models (LLMs). This approach applies the TEL (Test-Enhanced Learning) principle to mitigate the limitations of the conventional CPT+IT method, particularly in learning target domain knowledge and maintaining long-term memory. We introduce a CPT method utilizing descriptive QA and a strategy for efficiently constructing training data, which have demonstrated positive experimental results from a domain adaptation perspective. 
In the finance domain, \TELLME\ achieved up to a 23.6\% performance improvement on the finance benchmarks compared to existing methods. 

\section*{Limitations}
The \TELLME\ method proposed in this study has the following possible limitations.


\paragraph{Model Size.} Second, although we extended our study to a 70B‑parameter model~(\ref{appn:model_scaling}), the available computational budget required parameter‑efficient fine‑tuning, namely Low‑Rank Adaptation (LoRA) and 4‑bit quantization. These techniques reduce memory footprint and training time, but they also introduce additional variables, such as rank selection and quantization noise, that may interact with \TELLME. While the preliminary gains at this scale are encouraging, they may not faithfully represent \TELLME's effect on a fully dense 70B model. A systematic investigation without compression, and across even larger architectures, remains an important direction for future work.

\paragraph{Domain Diversity.}  Finally, this study focuses on finance and medicine, two domains known for their specialized and complex content. However, this scope does not cover the full range of real-world applications. Expanding \TELLME\ to additional domains would require reliable benchmark datasets and evaluation metrics, which are not always publicly available. We acknowledge this limitation and encourage further research to extend \TELLME\ to a broader range of domains, ideally alongside the development of standardized benchmarks in those areas.

\section*{Acknowledgement}
This work was supported by the affiliated institute of ETRI[2025-050] and Institute of Information \& communications Technology Planning \& Evaluation (IITP) grant, funded by the Korea government (MSIT) (No.RS-2024-00456709). We have used GPUs from High-Performance Research AI Computing Infrastructure Support at the 2 PFLOPS Scale (RS-2025-02653113)

\bibliography{custom}
\clearpage
\appendix

\section{Training Details and Hyperparameters}\label{Appendix:Training-Details-and-Hyperparameters}

\subsection{Training Setup} 

We use PyTorch as the primary deep learning framework, along with the HuggingFace Transformers library for efficient model training. The model is trained on a system equipped with eight NVIDIA A100 GPUs (80GB VRAM). Mixed-precision training with bfloat16 is enabled to reduce memory usage and improve computational efficiency.

The training process follows a single stage fine-tuning approach, where the model is initialized with a pre-trained checkpoint and adapted to the target domain using task-specific data. A cosine learning rate scheduler with a warm-up ratio of 0.03 is applied to prevent unstable updates in the early training phase.

\subsection{Training Datasets}
In this section, we summarize the characteristics of the two main domain datasets additionally utilized in this paper.

    
\paragraph{Bloomberg} This dataset, extracted from Bloomberg News, focuses on content that is relevant to the financial community. It provides documents of various lengths, offering domain specific terminology and real market trend information from the finance sector.

\paragraph{PubMed} Constructed from open-access data containing a large-scale collection of abstracts from the fields of medicine and life sciences, it enriches the medical domain with specialized knowledge, such as disease names, drug names, and clinical research terminologies that is often lacking in general language models, thereby promoting performance improvements in the respective field.

\subsection{Model Description}
 \paragraph{Llama-3.2-1B} An open-source model released by Meta, with 1 billion (1B) parameters. This lightweight model is designed for efficient performance with low computational cost, providing fundamental natural language understanding and task reasoning capabilities.

\paragraph{Llama-3.2-3B} An open-source model released by Meta, containing 3 billion (3B) parameters. It offers stronger contextual understanding and better generalization compared to the 1B model, making it more suitable for a variety of natural language processing (NLP) tasks.

\paragraph{Llama-3.1-8B} An open-source model released by Meta, equipped with 8 billion (8B) parameters. It is trained on diverse datasets, enabling strong natural language understanding and generation. It also excels in long-context understanding and domain adaptation.

\paragraph{SmolLM2-1.7B} An open-source model, developed by HuggingFaceTB, featuring 1.7 billion (1.7B) parameters. It is highly computationally efficient and, despite its smaller size, is designed to deliver strong performance in various natural language understanding tasks.

\subsection{Optimization and Training Strategy} 
We optimize the model using AdamW-8bit, a memory-efficient variant of AdamW, with a weight decay of 0.01 to prevent overfitting. The learning rate is set to 5e-5 and is gradually reduced following a cosine schedule. Training is performed with a batch size of 1, and gradient accumulation steps of 16 are used to achieve an effective batch size of 16. 

Since our focus is on continual pre-training, we limit the training process to 1 epoch to prevent catastrophic forgetting while allowing the model to adapt effectively to the target domain. In continual learning settings, over-training on new data can lead to the erosion of previously learned knowledge. By training for only one epoch, we ensure that the model retains its general knowledge while gradually adapting to domain-specific nuances. This approach aligns with previous findings in continual pre-training literature, where limited exposure to new data helps maintain a balance between adaptation and retention.

\subsection{Hyperparameter settings} 
\begin{table}[!th]
    \small
    \centering
    \begin{tabular}{ll}
        \midrule
        \textbf{Hyperparameter}      & \textbf{Value}  \\
        \midrule
        Data-type                   & bfloat16    \\
        Learning-rate               & 5e-5        \\
        Warm-up ratio               & 0.03        \\
        Learning-rate scheduler     & cosine      \\
        Optimizer                   & AdamW-8bit  \\
        Weight decay                & 0.01     \\
        Batch size                  & 1        \\
        Gradient accumulation steps & 16       \\
        Training epochs             & 1        \\
        \bottomrule
    \end{tabular}
    \caption{Hyperparameter settings used for training. This table summarizes the key hyperparameters, including learning rate, optimizer, batch size, and training schedule.}
    \label{tab:appendix-hyperparameter-settings}
\end{table}
    
\section{Evaluation Setup}\label{Appendix:Evaluation-setup}

\subsection{Experimental Domains and Selection of the QA Generation Model}

In this study, we constructed the \TELLMES dataset using the GPT4o-mini model based on our proposed prompt. Following prior research \citep{pezeshkpour2025learning, phasook2024thaibkd}, we chose medicine and finance as the target domains for the training data. We found that an open-source model (Llama3.3-70B) could also generate data of comparable quality. However, in terms of efficiency, renting GPUs to use the open-source language model was both more time-consuming and more expensive than employing GPT4o-mini. Consequently, we opted to use GPT for data generation.

\subsection{Evaluation Settings}
Essentially, all benchmarks were evaluated using accuracy as the primary metric in a 4-shot in-context setting. However, \rev{due to significant class imbalance in the FOMC and NIFTY datasets, the F1 score was employed as the evaluation metric in a zero-shot setting.}

\subsection{Finance Benchmarks}

\paragraph{FOMC (Federal Open Market Committee)}
This dataset consists of documents related to the Federal Open Market Committee (FOMC). It includes FOMC meeting minutes, press conferences, and speeches, and is used to evaluate the performance of models analyzing texts related to monetary policy and financial markets.

\paragraph{NIFTY (News-Informed Financial Trend Yield)}
This is a benchmark dataset constructed based on news and analytical materials related to the U.S. financial market, including news headlines from February 2019 to September 2020. It is used to assess models’ domain knowledge in areas such as the stock market, economic indicators, and corporate finance, as well as to measure their understanding and reasoning capabilities with respect to financial texts.

\paragraph{Massive Multitask Language Understanding - Finance (MMLU-F)}
MMLU for the finance domain is a benchmark extracted from a subset of the Massive Multitask Language Understanding (MMLU) dataset, specifically focused on finance-related disciplines. This benchmark was extracted by selecting a diverse set of subjects relevant to financial studies, including business ethics, econometrics, high school macroeconomics, high school microeconomics, management, marketing, and professional accounting. 

\subsection{Medicine Benchmarks}

\paragraph{HEAD-QA (HEAlthcare Dataset)}
HeadQA is a multiple-choice question-answering benchmark designed to advance research in complex reasoning. The dataset consists of questions taken from exams required for specialized roles in the Spanish healthcare system, posing significant challenges even for experts in the field.

\paragraph{MedMCQA}
A large-scale Multiple-Choice Question Answering (MCQA) dataset created to tackle real-world medical entrance exam questions. The MedMCQA task can be defined as $X = \{Q, O\}$, where $Q$ denotes the textual questions and $O$ represents the set of possible answer choices. Each question is accompanied by multiple candidate answers, $O = \{O1, O2, …, On\}$, and the objective is to identify the correct single or multiple answers from the given options.

\paragraph{MMLU-C (Massive Multitask Language Understanding - Clinical)}
MMLU-Clinic is a part of the MMLU benchmark that includes multiple-choice questions related to the medical field, designed to assess the medical knowledge understanding of large language models.
This dataset covers various medical subfields such as anatomy, genetics, and clinical knowledge, and is used to evaluate models like Med-PaLM 2.
Additionally, it is utilized alongside MedQA, MedMCQA, and PubMedQA to assess the precision of LLMs in reasoning and answering questions in the medical domain.
Notably, it can also be applied to professional evaluations like medical licensing exams, making it a significant benchmark in AI research for the healthcare sector.

\section{Detailed Description for \instpt\ Dataset Generation}\label{Appendix:Detailed-InstPT}
For the construction of \instpt\ data, we built the training dataset based on the code provided by \citet{cheng-etal-2024-instruction}\footnote{https://github.com/microsoft/LMOps}. Specifically, for \instpt\ QA generation, we utilized the open-source synthesizer based on Mistral 7B~\citep{jiang2023mistral} released by the authors\footnote{https://huggingface.co/instruction-pretrain/instruction-synthesizer}. The parameters used for data generation were set with max\_new\_tokens as 2048, conducted in a 3-shot setting.


\section{\TELLMES Dataset Examples and Clarification}\label{Appendix:Tellme-Dataset-Examples-and-Clarification}

\subsection{\TELLMES Dataset Examples}

Table~\ref{appendix:tellme_example_1}, \ref{appendix:tellme_example_2} illustrate examples of datasets generated using the \TELLME\ approach in the Medicine and Finance domains. The QA pairs are related to the text but are designed to introduce new knowledge that cannot be directly found within the given text.

\subsection{Comparison example with the \instpt\ dataset} 
Table~\ref{appendix:tab_tellme_instpt} presents examples of QA datasets generated using the \TELLMES approach and the \instpt\ approach for the same plain text.

\subsection{Plain text dependency of the TELLME QA dataset}
To assess the extent to which QA datasets rely on plain text, we define the Coverage Ratio (CR). Equation~\ref{equ:coverage-ratio} presents the formula used to compute CR, which is calculated as the proportion of words($w$) in the answer that also appear in the corpus, relative to the total number of words in the answer. 
\begin{equation}
    \textbf{CR} = \frac{\text{length}(\{w \mid w \in \mathbf{a} \cap w \in \mathbf{t} \})}{\text{length}(\{w \mid w \in \mathbf{a}\})} * 100   
\label{equ:coverage-ratio}
\end{equation}
The experiment was conducted using a training corpus from the finance domain, comparing QA datasets generated using the \TELLMES approach and the \instpt\ approach. 
\begin{table}[!h]
\centering
\small
\begin{tabular}{l c c c c}
    \toprule
    \multirow{2}{*}{\textbf{Domain}} & \multicolumn{2}{c}{\textbf{Finance}} & \multicolumn{2}{c}{\textbf{Medical}} \\
    \cmidrule(lr){2-3} \cmidrule(lr){4-5}
    & \textbf{Basic} & \textbf{w/o stop} & \textbf{Basic} & \textbf{w/o stop} \\
    \midrule
    \TELLME  & 32.35 & 14.83 & 48.70 & 34.82 \\
    \instpt & 87.14 & 86.60 & 69.98 & 61.31 \\
    \bottomrule
\end{tabular}
\caption{‘Basic’ refers to the model-generated answer, while ‘w/o stop’ refers to the answer with stopwords removed.}
\label{tab:coverage_ratio}
\end{table}

Table~\ref{tab:coverage_ratio} shows that the dataset constructed using the \TELLMES approach has a significantly lower CR compared to the \instpt\ approach. This suggests that the \TELLMES dataset does not rely solely on plain text but also requires external knowledge beyond the given text.

\subsection{Data Composition and Loss Function Design}\label{appendix:d-4}
\begin{figure*}[t]
  \includegraphics[width=\linewidth]{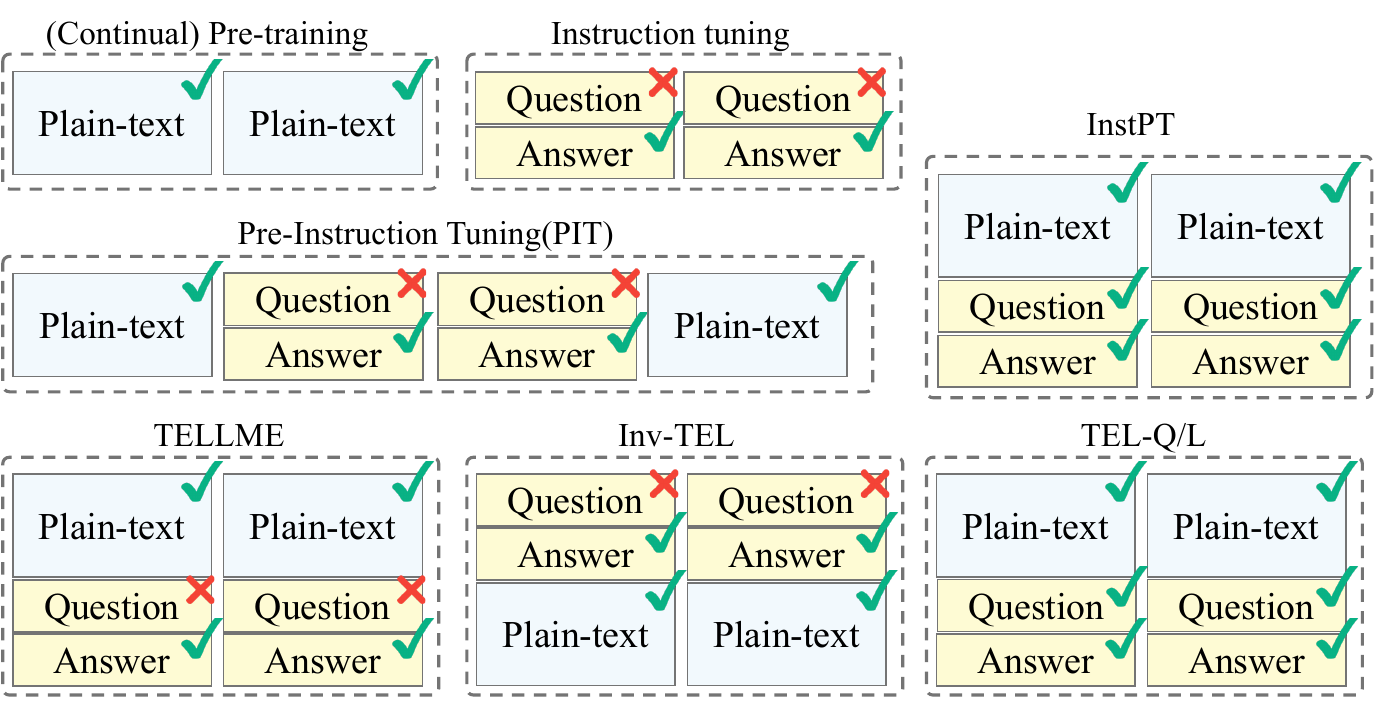}
    \caption{Comparison of data composition and token-level loss masking across training paradigms.
Dashed boxes denote individual mini-batches; within each, the vertically stacked panels constitute a single training sequence. (Continual) Pre-training consumes plain text only, applying the objective to every token. Instruction tuning feeds question–answer (QA) pairs but back-propagates loss exclusively on answer tokens. Pre-Instruction Tuning (PIT) interleaves plain text with QA pairs, computing loss on plain-text and answer tokens while masking question tokens. InstPT activates the objective for all tokens in both plain-text and QA examples. Our proposed TELLME keeps PIT’s masking strategy while doubling the proportion of plain-text sequences to reinforce language modeling; its counterpart Inv-TEL inverts the ordering of QA and plain-text segments within each sample. TEL-Q/L serves as a fully supervised upper bound, applying loss to all tokens in both modalities. Green check marks identify tokens whose losses are calculated, whereas red Crossmark denote tokens excluded from loss computation.}
  \label{fig:appendix_loss_variation}
\end{figure*}

Figure~\ref{fig:appendix_loss_variation} provides an overview of the data composition and loss calculation strategies employed in the training of various methods, including CPT, IT, PIT, InstPT, \TELLME, Inv-TEL, and TEL-Q/L.

In this figure, plain-text refers to the corpus typically used for domain adaptation, while question and answer denote the components of QA datasets. These data types can be treated either as independent samples or concatenated into a single sequence during training. For instance, PIT adopts the former strategy, whereas methods such as \TELLME and InstPT follow the latter. In concatenated settings, the autoregressive nature of language models makes the relative ordering between plain-text and QA data a critical factor influencing learning outcomes.

In Figure~\ref{fig:appendix_loss_variation}, green check marks indicate the positions where the loss is applied, whereas crossmark denote regions excluded from the loss calculation. It is common in instruction tuning setups to exclude the question portion of QA data from loss computation. However, several CPT-integrated QA approaches (e.g., InstPT) include the question in the prediction objective. To investigate the impact of this design choice, we additionally evaluate a variant, TEL-Q/L, in which the question loss is explicitly excluded during training.

\begin{table*}
\begin{minipage}{0.95\textwidth}
    \begin{tabular}{|>{\arraybackslash}p{\textwidth}|}
        \hline
        \rowcolor{gray!20}
        \begin{minipage}{\textwidth}
            \centering
            \textbf{Plain Text}
        \end{minipage}\\
        \hline
        {\scriptsize U.K. Pound Weakens Versus Dollar Before BOE Bond-Purchase, Rate Decision.
        (…)
        The Monetary Policy Committee will leave its key rate unchanged at 0.5 percent, according to all 53 economists surveyed by Bloomberg. Eleven of 32 economists surveyed separately predict at least a 50 billion-pound (\$77 billion) increase in its so- called quantitative easing program. “There might be a short relief rally in sterling on a lack of further BOE action but it’s likely to be short- lived as the market is still focused on the likelihood of further QE by the November meeting,” said Ian Stannard , London-based head of European currency strategy at Morgan Stanley. “The BOE will move towards further QE in coming months and ultimately sterling will come under pressure.” The pound weakened 0.1 percent to \$1.5439 at 8:54 a.m. in London. It declined 0.2 percent to 118.47 yen and was little changed against the euro at 86.36 pence. Morgan Stanley recommends selling the pound on any “relief rallies” that are likely to occur should the central bank refrain from announcing further monetary stimulus for the U.K. economy. The Bank of England has faced pressure to embark on further quantitative easing to help revive an economy battling the steepest government spending cuts since World War II and a worsening euro-area debt crisis. The U.K.’s economy} \\
        \hline
        \rowcolor{gray!20}
        \begin{minipage}{\textwidth}
            \centering
            \textbf{QA from \TELLME}
        \end{minipage}\\

        \begin{minipage}[t]{\textwidth}
            {\scriptsize
                Q1 : What is the purpose of a central bank's quantitative easing program?\\
                A1 : The purpose of a central bank's quantitative easing program is to stimulate the economy by increasing the money supply, lowering interest rates, and encouraging lending and investment during periods of economic downturn.\\
                Q2 : How can currency exchange rates be affected by a country's monetary policy decisions?\\
                A2 : Currency exchange rates can be affected by a country's monetary policy decisions as changes in interest rates or quantitative easing can influence investor sentiment, capital flows, and overall demand for the currency, leading to fluctuations in its value against other currencies.\\
                Q3 : What are the potential risks of investing in government bonds during times of economic uncertainty?\\
                A3 : The potential risks of investing in government bonds during times of economic uncertainty include interest rate risk, where rising rates can decrease bond prices, inflation risk, where the purchasing power of returns may be eroded, and credit risk, where concerns about the government's ability to meet its obligations could lead to defaults.}\\
        \end{minipage} \\
        \hline
    \end{tabular}
\end{minipage}
\caption{An example of a QA dataset generated using the \TELLME\ approach in the finance domain.}
\label{appendix:tellme_example_1}
\end{table*}
\begin{table*}
\begin{minipage}{0.95\textwidth}
    \begin{tabular}{|>{\arraybackslash}p{\textwidth}|}
        \hline
        \rowcolor{gray!20}
        \begin{minipage}{\textwidth}
            \centering
            \textbf{Plain Text}
        \end{minipage}\\
        \hline
        {\scriptsize Low-Energy Hawkins Type III Talar Neck Fracture-Dislocation With Neurovascular and Tendon Entrapment in a Pediatric Patient. Several serious complications can occur after talar neck fractures. However, these fractures are extremely rare in children. We present a pediatric low-energy Hawkins type III fracture-dislocation that had excessive displacement accompanied by neurovascular and tendon entrapment. A 9-year-old male patient referred to our hospital 5 hours after jumping off a swing in a children's playground.An excessively displaced talar neck fracture-dislocation was observed at the initial evaluation. The patient underwent urgent surgery. The tibialis posterior flexor digitorum longus tendons, posterior tibial artery, and tibial nerve were entrapped at the fracture site. The talar neck fracture was reduced using open reduction. The neurovascular structures and tendons were removed from the fracture site. The fracture was fixed using two 4.5-mm cannulated screws. The patient was able to bear full weight at 10 weeks postoperatively. At 6 months, the patient was able to walk unassisted with full ankle range of motion. However, at 2 years, his American Orthopaedic Foot and Ankle Society Ankle-Hindfoot scale score had decreased to 72 points, and we observed avascular necrosis in the talar head (…) In the pediatric population, even low-energy trauma, such as had occurred in our patient, can result in severe displaced fracture dislocations. After severe displaced fracture-dislocations, important soft tissue structures can become entrapped between fracture fragments, and surgeons should be aware of this situation when considering using closed reduction.} \\
        \hline
        \rowcolor{gray!20}
        \begin{minipage}{\textwidth}
            \centering
            \textbf{QA from \TELLME}
        \end{minipage}\\

        \begin{minipage}[t]{\textwidth}
            {\scriptsize
                Q1 : What are some common complications associated with talar neck fractures?\\
                A1 : Common complications include avascular necrosis, neurovascular injury, and tendon entrapment.\\
                Q2 : Why are talar neck fractures considered rare in the pediatric population?\\
                A2 : Talar neck fractures are rare in children due to the relative strength and flexibility of pediatric bones compared to adults.\\
                Q3 : What is the typical treatment approach for severely displaced talar neck fractures?\\
                A3 : The typical treatment involves surgical intervention, often requiring open reduction and internal fixation}\\
        \end{minipage} \\
        \hline
    \end{tabular}
\end{minipage}
\caption{An example of a QA dataset generated using the \TELLME\ approach in the medicine domain.}
\label{appendix:tellme_example_2}
\end{table*}

\clearpage

\newpage

\clearpage

\begin{table*}
\begin{minipage}{0.95\textwidth}
    \begin{tabular}{|>{\arraybackslash}p{\textwidth}|}
        \hline
        \rowcolor{gray!20}
        \begin{minipage}{\textwidth}
            \centering
            \textbf{Plain Text}
        \end{minipage}\\
        \hline
        {\scriptsize China's Faster Inflation Fuels Speculation Rate-Rise Near.China’s central bank may \TELLMEBG{raise interest rates} within weeks after \TELLMEBG{inflation} accelerated to the fastest pace in 25 months in October, a Bloomberg News survey of economists showed. The benchmark one-year lending rate will rise to \instptBG{5.81 percent} by year-end from \instptBG{5.56 percent,} according to the median forecast of 11 analysts polled after yesterday’s price data. The deposit rate may climb to \instptBG{2.75 percent} from \instptBG{2.5 percent,} the survey showed. China’s benchmark Shanghai Composite Index slid 2.6 percent as of 1:01 p.m. local time on speculation that officials may move as early as today or this weekend after increasing banks’ \TELLMEBG{reserve requirements} on Nov. 10. Higher rates could complicate government efforts to limit gains in consumer and property prices by luring more money to the fastest-growing major economy. “We know there’ll be more tightening given how inflation has accelerated and home prices haven’t come down, but the sudden talk that there may be an interest-rate hike as early as the end of today really spooked the markets,” said Mark Tan , who helps oversee \$12 billion at UOB Asset Management Ltd. Goldman Sachs Group Inc. said yesterday that October’s economic data indicated Chinese growth was “firm” and more “policy tightening” is needed. Industrial \& Commercial Bank of China Ltd. and China Vanke Co. led lenders and property developers lower as the benchmark index fell the most in three months. Reserve Requirements Price pressures in China’s economy may be exacerbated by the nation’s currency curbs and imbalances in trade and capital flows that Group of 20 leaders are meeting in Seoul to tackle. The central bank announced a 0.5 percentage point increase in lenders’ reserve requirements this week after the customs bureau reported that October’s trade surplus surged to \$27.1 billion. Some lenders including Bank of Communications Co., had an additional increase, meaning their requirements rose a full percentage point, two people with direct knowledge of the situation said yesterday. Brian Jackson , an emerging markets strategist at Royal Bank of Canada in Hong Kong, said yesterday that more interest-rate increases are “clearly on the way.” The central bank raised rates last month for the first time since 2007 as part of an exit from crisis policies that included scrapping in June the yuan’s peg to the dollar. A record expansion in lending has added to concern that China’s inflation, now centered on food costs, will broaden. In October, new lending was a more-than-forecast \instptBG{587.7 billion yuan} (\$\instptBG{89 billion)}, a central bank report showed yesterday. While the latest inflation number was above all 28 estimates in a Bloomberg News survey of economists, it matched rumors cited before the data by firms including Guoyuan Securities Co. in Shanghai. \instptBG{``There's talk of an interest-rate hike over the weekend,''}
        said \instptBG{Wu Kan}, a Shanghai-based fund manager at Dazhong Insurance Co., which oversees \$285 million. “It’s quite possible given how inflation has accelerated.”} \\
        \hline
        \rowcolor{gray!20}
        \begin{minipage}{0.48\textwidth}
            \centering
            \textbf{QA from \TELLME}
        \end{minipage}%
        \begin{minipage}{0.5\textwidth}
            \centering
            \textbf{QA from \instpt}
        \end{minipage} \\
        \hline
        \begin{minipage}[t]{0.48\textwidth}
        \vspace{0.1cm}
            {\scriptsize
            Q1: What are the potential effects of \TELLMEBG{raising interest rates} on consumer spending and borrowing? \\
            A1: Raising interest rates typically leads to higher borrowing costs, which can reduce consumer spending and borrowing. As loans become more expensive, consumers may delay or reduce purchases, particularly for big-ticket items that often require financing, such as homes and cars.\\\\
            Q2: How can \TELLMEBG{inflation impact} the overall economy?\\
            A2: Inflation can erode purchasing power, leading to decreased consumer confidence and spending. If inflation rises too quickly, it may prompt central banks to increase interest rates to stabilize prices, which can further slow economic growth. Additionally,
            high inflation can disproportionately affect lower-income households, as they spend a larger portion of their income on essential goods and services.\\\\
            Q3: What is the role of \TELLMEBG{reserve requirements} in monetary policy?\\
            A3: Reserve requirements are regulations set by central banks that dictate the minimum amount of reserves a bank must hold against deposits. By increasing reserve requirements, a central bank can reduce the amount of money available for lending, thus tightening monetary policy. This can help control inflation but may also slow down economic growth by limiting credit availability.\\
            }
        \end{minipage}%
        \hspace{0.04\textwidth}
        \begin{minipage}[t]{0.48\textwidth}
        \vspace{0.1cm}
            {\scriptsize
             Q1: What is the name of the fund manager?\\
             A1: \instptBG{Wu Kan}\\
             Q2: What did he say?\\
             A2: \instptBG{``There's talk of an interest-rate hike over the weekend,''}\\
             Q3: What is the rate expected to rise to by the end of the year?\\
             A3: \instptBG{5.81 percent}\\
             Q4: What is the current rate?\\
             A4: \instptBG{5.56 percent,}\\
             Q5: What is the amount expected to be lent?\\
             A5: \instptBG{587.7 billion yuan}\\
             Q6: What is that converted to the thousands?\\
             A6: \instptBG{89 billion}\\
             Q7: What is the deposit rate expected to rise to?\\
             A7: \instptBG{2.75 percent}\\
             Q8: From what?\\
             A8:\instptBG{2.5 percent,}\\
             Q9: What is the net change?\\
             A9: \instptBG{0.25 percent,}\\
             Q10: What is the one year lending rate expected to be?\\
             A10: \instptBG{5.81 percent}\\
             Q11: What was it last year?\\
             A11: \instptBG{5.56 percent}\\}
        \end{minipage} \\
        \hline
    \end{tabular}
\end{minipage}
\caption{An example of QA datasets generated using the \TELLME approach and the InstPT approach for the same plain text. InstPT follows a reading-comprehension format, where answers are typically extractive. In contrast, \TELLME generates open-ended questions that require a deeper understanding of the text beyond surface-level extraction.}
\label{appendix:tab_tellme_instpt}
\end{table*}

\newpage
\begin{table*}[t]
    \centering
    \resizebox{\textwidth}{!}{
    \begin{tabular}{|p{0.95\textwidth}|}
    \hline
    \rowcolor{gray!20}
    \textbf{System message} \\
    \hline
    \vspace{0.005cm}
    You are a {\color{red}{\{domain\}}} Q\&A generator. You will be provided with a {\color{red}{\{domain\}}} context excerpt, but the solver does NOT see it. Therefore, you must: \\

    \begin{enumerate}
        \item Avoid direct questions about any specific events or data in the excerpt.
        \item Instead, create questions based on general {\color{red}{\{domain\}}} knowledge.
        \item Ensure each question can be answered independently of the excerpt, since the solver does not have access to it.
        \item Provide exactly three open-ended question-answer pairs in English.
        \item Output your response strictly in JSON format with no additional explanation.
    \end{enumerate}

    Output format requirements: \\
    - Create an array named ``questions\_and\_answers''. \\
    - For each Q\&A pair, provide an object with the keys ``question'' and ``answer''. \\
    - Do not include any text outside the JSON structure. \\\\
    \hline
    \rowcolor{gray!20}
    \textbf{User message} \\
    \hline
    Here is the {\color{red}{\{domain\}}} context excerpt for your reference:\\\\
    
    \noindent\{\textsc{Input text}\}\\\\
    
    Please produce exactly three {\color{red}{\{domain\}}}-related question-answer pairs in the specified JSON format, without referencing specific details from the text and without adding extra commentary.\\
    \hline
    \end{tabular}
    }
    \caption{Prompt for \TELLME\ Dataset Construction. In this study, \{domain\} refers to either ``medicine'' or ``finance,'' depending on the context. The system message defines the task, while the user message provides the article excerpt to guide question generation. The \{\textsc{Input text}\} corresponds to a PubMed article excerpt for the ``medicine'' domain and a Bloomberg article excerpt for the ``finance'' domain.}
    \label{tab:Appendix-GPT-Prompt}
\end{table*}

\clearpage

\section{Detailed Methodology for TEL Dataset Construction}\label{Appendix:TEL-Dataset-Info}
In this study, we constructed a TEL dataset using GPT-generated content tailored to different domains. Specifically, we designed structured prompts to generate high-quality question-answer (QA) pairs in the medical and finance fields. These prompts were crafted to ensure the generated questions were independent of specific article excerpts while remaining relevant to the broader domain knowledge.

\subsection{Description of the Prompt Design for Domain Specific Dataset}

As shown in Table~\ref{tab:Appendix-GPT-Prompt}, the prompt design for TEL dataset construction includes a system message and a user message. The system message defines the task for generating domain-specific Q\&A pairs, while the user message provides the article excerpt as input. The \{\textsc{Input text}\} in the user message is sourced from PubMed for the ``medicine'' domain and Bloomberg for the ``finance'' domain. This structured prompt ensures the generation of high-quality Q\&A pairs that are independent of specific details in the provided excerpts.

\subsection{Ensuring Context Isolation: Filtering for QA Dataset}
To construct the \TELLMES dataset, we designed a prompt using GPT-4o-mini to generate Question \& Answer pairs that can be solved without plain text. However, after reviewing 100k samples, we found that approximately 80 samples (0.008\%) contained keywords such as “in this context” and “described,” indicating that some questions and answers were generated in a way that required context. Although the number of such samples was small, this issue could compromise the fair evaluation of the \CPTSFT\ and \TELLMES methods. Therefore, we filtered out these samples before finalizing the \TELLME\ dataset.

\section{Expanding LLM Domains through Continual Learning
}\label{Appendix:case studies}

\paragraph{FINDAP: A Structured Approach for Financial LLM Adaptation~\citep{ke2025demystifying}.}
FINDAP applied a training methodology consisting of Financial-based Continual Pre-training, Instruction Tuning, and Preference Alignment to train a finance-specialized LLM. The PA (Preference Alignment) stage incorporates techniques proposed in the paper to enhance financial reasoning performance by introducing two methods: Stepwise Corrective Preference (SCP) and Final Answer Preference (FAP). SCP provides feedback by comparing the model’s reasoning process at each intermediate step with the correct answer, ensuring accurate step-by-step inference in financial problem-solving. Meanwhile, FAP guides the model to prefer more reliable answers when selecting the final response.

\paragraph{Swallow: Cross-Lingual Continual Pre-Training for Japanese LLMs~\citep{fujii2024continual}.}
Swallow is a study that applied Cross-Lingual Continual Pre-Training to enhance Japanese language performance. This research analyzes the impact of vocabulary expansion and the use of parallel corpora in the process of adapting an English-centric LLM to Japanese.

The training process of the Swallow model followed three stages:
(1) Continual Pre-Training using a Japanese corpus,
(2) Additional training with a Japanese-English parallel corpus,
(3) Application of Japanese-specific vocabulary expansion.

Through this approach, the model effectively improved English-Japanese machine translation performance.

\paragraph{Don’t Stop Pretraining: Adapt Language Models to Domains and Tasks~\citep{gururangan-etal-2020-dont}.}
This study proposes Domain-Adaptive Pretraining (DAPT) and Task-Adaptive Pretraining (TAPT) to enhance the performance of large language models (LLMs). DAPT strengthens domain adaptation by further training the model on large-scale data from a specific domain, while TAPT improves task performance by additional training on task-specific data.

Experimental results show that applying both DAPT and TAPT together yields the highest performance, while in certain tasks, TAPT alone is sufficient for significant improvement. This suggests that an appropriate additional training strategy is more effective than merely increasing model size. Therefore, the study emphasizes the importance of tailored training strategies for domain- and task-specific optimization in NLP models.

\paragraph{Efficient Continual Pre-training for Building Domain-Specific Large Language Models.~\citep{su-etal-2023-efficient}}
This study proposes Continual Pre-training (CPT) as a cost-effective way to build domain-specialized Large Language Models (LLMs). By developing the FinPythia model in finance and applying DACP and TACP, performance improved by up to 8.3\%.

Furthermore, selecting only key data (ETS-DACP, ETA-DACP) instead of full dataset training cut costs by 90\% while maintaining performance. Despite domain-specific gains, open-domain performance remained stable, proving the method’s broad applicability.

The study emphasizes that CPT is a more practical and economical alternative to training LLMs from scratch.

\section{\rev{Further Analysis of the \TELLME}}\label{further-anal}
\subsection{Cost-efficient Training of \TELLME}
\begin{figure}[!h]
    \centering
    \includegraphics[width=\linewidth]{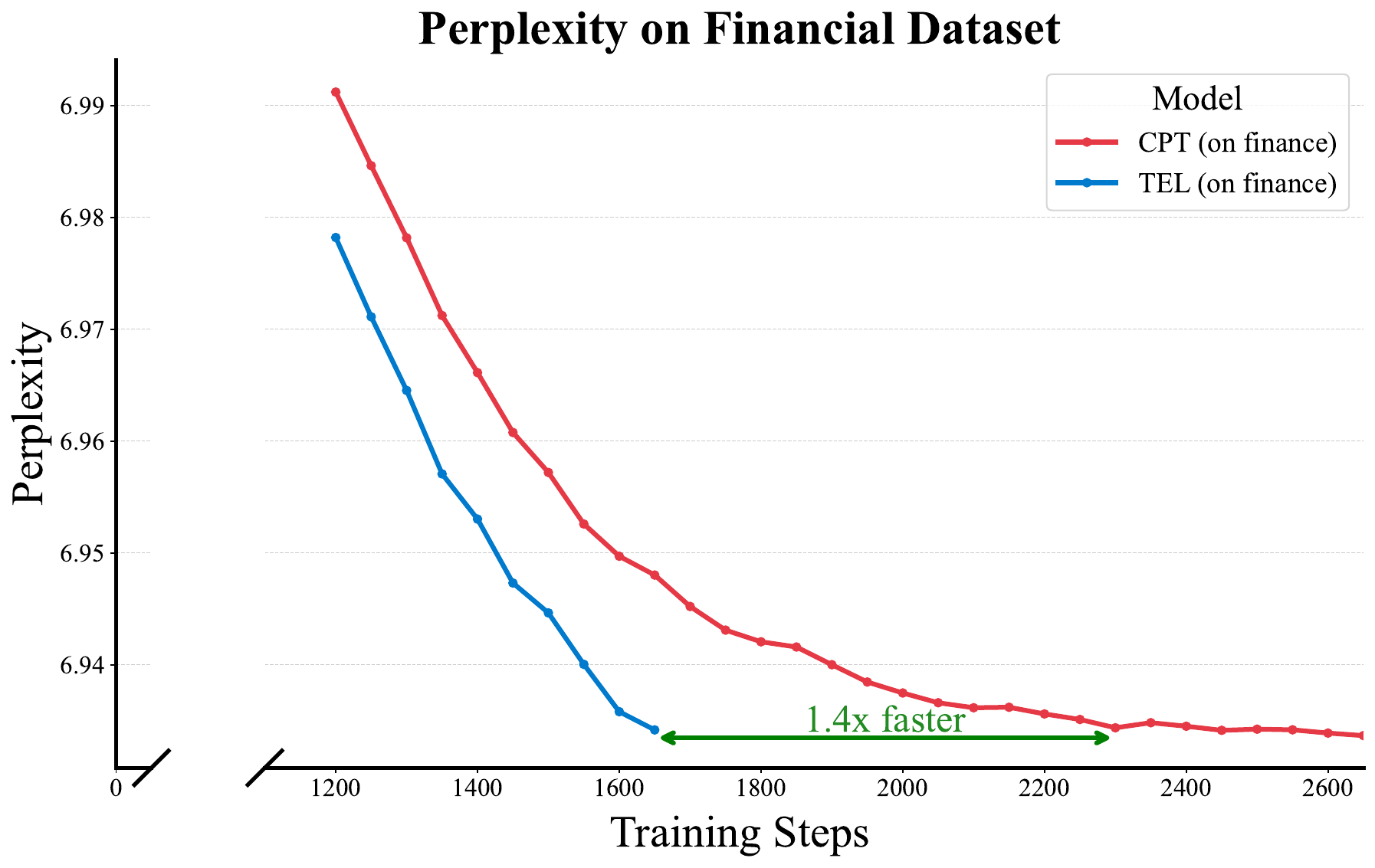}
    \caption{Perplexity of CPT and \TELLME\ methods based on training steps.}
    \label{fig:ppl-cost-effect}
\end{figure}

Figure~\ref{fig:ppl-cost-effect} illustrates the PPL scores of the CPT and \TELLME\ methods over the training steps in the finance domain based on SmolLM2-1.7B. The \TELLME-based model achieved a PPL of approximately 6.935 after 1,650 steps, whereas the CPT-based model reached a similar level only after 2,300 steps. This indicates that the CPT model requires approximately 1.4 times more training time to achieve the same performance as the TEL model. Consequently, the \TELLME\ method demonstrates its cost efficiency by achieving superior performance within a shorter training duration.

\subsection{Analysis of the \TELLMES Indicator on Different Datasets}
\begin{table}[!h]
\centering
\tiny
\begin{tabular}{lcccc}
\toprule
Model                                         & FOMC     & NIFTY    & MMLU-F   & AVG      \\
\midrule
Llama-3.2-1B                    \\
$\quad$ + \instpt                             & 28.75    & 33.38    & 45.73    & 35.95    \\
\rowcolor{gray!20} 
$\quad$ + $\indicator(x_i \in \mathbf{q}) = 0$      & 27.61    & 35.51    & 46.70    & 36.61    \\
\midrule

Llama-3.2-3B                    \\
$\quad$ + \instpt                             & 28.79    & 19.60    & 44.68    & 31.03     \\
\rowcolor{gray!20} 
$\quad$ + $\indicator(x_i \in \mathbf{q}) = 0$      & 32.44    & 25.86    & 45.98    & 34.76    \\
\midrule

Llama-3.1-8B                    \\
$\quad$ + \instpt                             & 34.40    & 27.30    & 49.26    & 36.99        \\
\rowcolor{gray!20} 
$\quad$ + $\indicator(x_i \in \mathbf{q}) = 0$      & 38.97    & 30.35    & 51.40    & 40.24    \\
\midrule

SmolLM2-1.7B                    \\
$\quad$ + \instpt                             & 28.75    & 33.38    & 45.73    & 35.95           \\
\rowcolor{gray!20} 
$\quad$ + $\indicator(x_i \in \mathbf{q}) = 0$      & 27.61    & 35.51    & 46.70    & 36.61    \\
\bottomrule
\end{tabular}
\caption{Comparison of model performance on the \instpt\ dataset when optimized with and without the indicator proposed by \TELLME. Here, \instpt\ refers to the method using both the dataset and the indicator proposed by \instpt\, while $\indicator(x_i \in \mathbf{q}) = 0$ denotes the method that utilizes the dataset proposed by \instpt\ but applies the indicator proposed in this study.}
\label{tab:appendix_instpt_tel}
\end{table}

Table~\ref{tab:appendix_instpt_tel} presents a performance comparison when optimizing the \instpt\ dataset using the indicator proposed by \TELLME. The experimental results show that employing \TELLME's indicator consistently led to superior performance across all models. Furthermore, considering the results in Table~\ref{tab:anl-table}, where \TELLMES outperformed TEL-Q/L, these findings suggest that incorporating the proposed indicator in the optimization process for QA-based Continual Learning dataset can be more effective.

\subsection{Performance Variation Based on the Proportion of the \TELLMES Dataset.}
\label{app:dataset-quality}
\begin{figure}[!h]
    \centering
    \includegraphics[width=\linewidth]{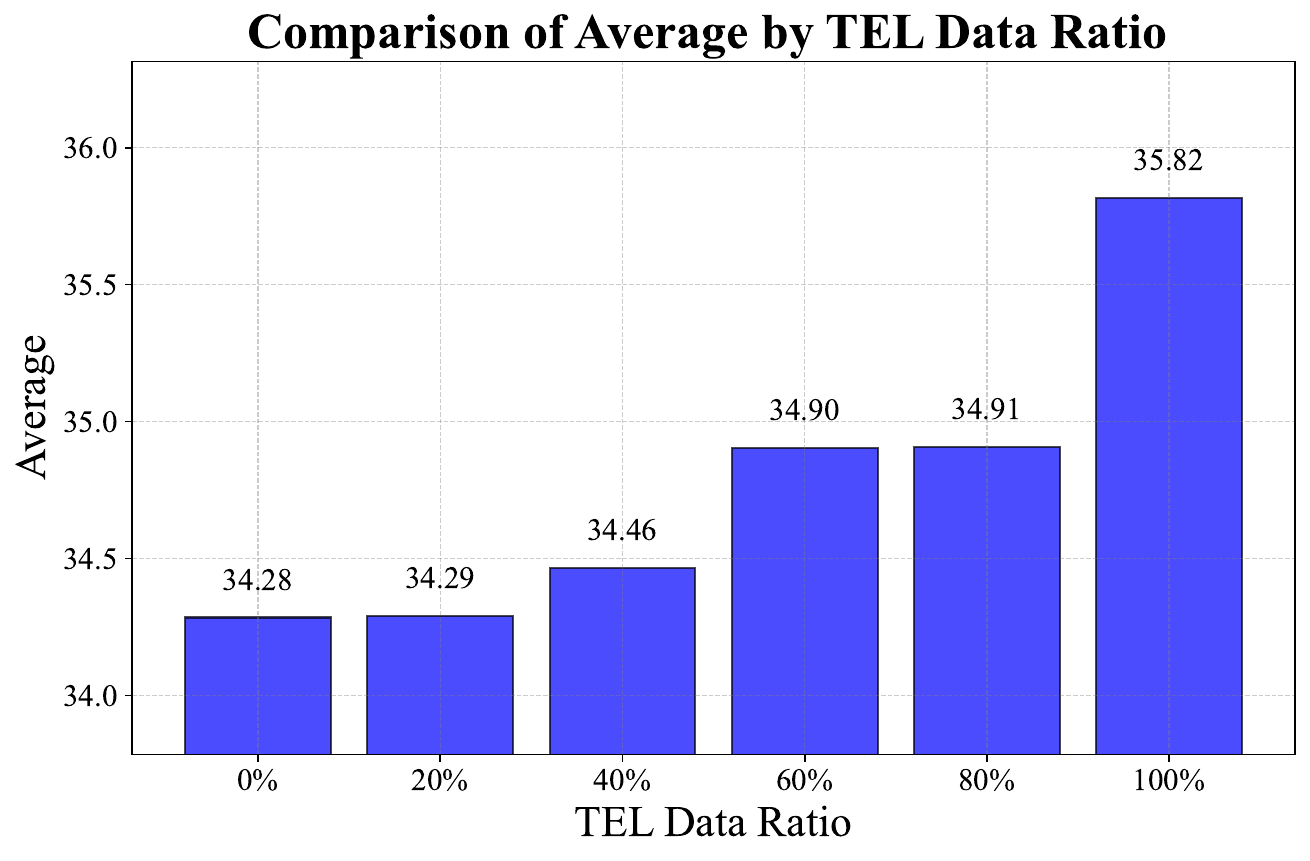}
    \caption{Performance chart illustrating the effect of different ratios of general plain text and the \TELLME\ dataset, using a model trained on finance data. The evaluation is the same financial benchmark used in Table~\ref{tab:main_table}}
    \label{fig:raio}
\end{figure}

\begin{figure}[!h]
    \centering
    \includegraphics[width=\linewidth]{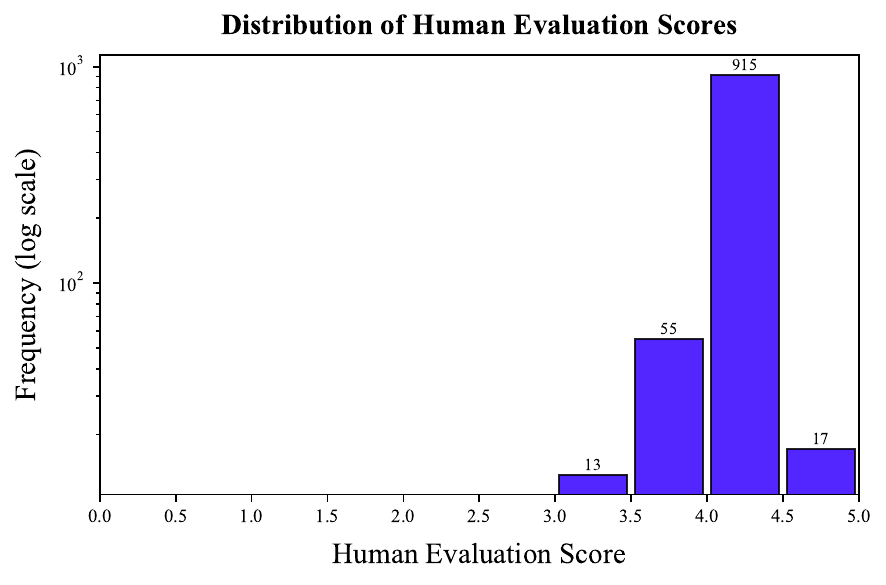}
    \caption{Distribution of averaged human evaluation scores for the finance domain.
    Each data point represents the mean of human-evaluated quality scores across three question–answer pairs associated with a given financial text.
    The scores are aggregated on a 1–5 scale, where higher values indicate better factuality, coherence, and domain correctness.
    Notably, the overall distribution is concentrated above a score of 3, reflecting the high linguistic and conceptual quality of the curated financial dataset.}
    \label{fig:human_eval}
\end{figure}

Figure~\ref{fig:raio} illustrates the impact of the dataset composition ratio between CPT and TEL datasets on model performance when training the SmolLM2-1.7B model in the Finance domain using a 100k dataset. As shown in the figure, model performance tends to improve as the proportion of the TEL dataset increases.
When the TEL dataset comprises 20\% of the total data, the model's performance is comparable to that of a model trained exclusively on the CPT dataset. However, when the TEL dataset ratio increases to 40\%, the model achieves approximately 0.18 points higher performance than the CPT-based model. Additionally, when TEL accounts for 60\% or 80\% of the dataset, the model's performance remains nearly identical at around 34.9 points, marking an improvement of approximately 0.62 points over the CPT-based model.
At a TEL dataset ratio of 100\%, the model achieves its highest performance, reaching a score of 35.82. Notably, even with only 60\% TEL data, the model exhibits significant performance gains while maintaining a relatively low training cost of approximately \$7.2, making it a cost-effective choice.

In addition, we evaluated the quality of the generated data. Figure~\ref{fig:human_eval} presents the results of the LLM-as-a-judge evaluation conducted using GPT-4 mini, based on the following automatic evaluation prompt: \textit{Please evaluate the quality of the following question and answer pair based on relevance, clarity, and completeness. Provide a single quality score between 1 (poor) and 5 (excellent).}
We performed the evaluation on 1,000 samples, and the dataset achieved an average score of 4.03, confirming the high quality of the constructed corpus.

\subsection{Scalability of \TELLMES Across Models}
\begin{table}[!h]
\centering
\tiny
\begin{tabular}{lcccc}
\toprule
Model                  & HeadQA     & MedMCQA    & MMLU-C   & AVG      \\
\midrule
gpt2-medium            & 24.36    & 21.90    & 26.60    & 24.29     \\
$\quad$ + \instpt      & 23.38    & 22.23    & 27.34    & 24.32     \\
\rowcolor{gray!20}
$\quad$ + \TELLME      & 24.65    & 22.54    & 28.66    & \textbf{25.29}     \\

\midrule
gpt2-xl                & 25.38    & 22.09    & 27.61    & 25.03     \\
$\quad$ + \instpt      & 25.20    & 22.45    & 28.98    & 25.54     \\
\rowcolor{gray!20}
$\quad$ + \TELLME      & 26.70    & 23.14    & 29.98    & \textbf{26.60}     \\

\midrule
Qwen2.5-0.5B           & 28.74    & 25.01    & 30.75    & 28.17     \\
$\quad$ + \instpt      & 27.10    & 24.10    & 31.93    & 27.71     \\
\rowcolor{gray!20}
$\quad$ + \TELLME      & 29.25    & 25.46    & 31.45    & \textbf{28.72}     \\

\midrule
Qwen2.5-3B             & 38.55    & 32.27    & 44.16    & 38.33      \\
$\quad$ + \instpt      & 35.30    & 30.41    & 42.15    & 35.95     \\
\rowcolor{gray!20}
$\quad$ + \TELLME      & 39.24    & 31.70    & 44.33    & \textbf{38.42}     \\

\midrule
phi-1\_5               & 27.90    & 24.53    & 35.41    & 29.28     \\
$\quad$ + \instpt      & 27.83    & 23.93    & 33.90    & 28.55     \\
\rowcolor{gray!20}
$\quad$ + \TELLME      & 28.70    & 24.65    & 34.71    & \textbf{29.35}     \\
\bottomrule
\end{tabular}
\caption{Performance comparison of various models optimized with \TELLMES on medicine benchmarks.}
\label{tab:appendix_diversity}
\end{table}

Table~\ref{tab:appendix_diversity} presents the performance of various models optimized with \TELLME. The results indicate that all models incorporating \TELLME\ exhibit improved average performance on medicine benchmark datasets. This suggests that \TELLME\ can be applied to a wide range of language models and effectively enhances performance.

\subsection{Scalability of \TELLMES Across Model size}\label{appn:model_scaling}
\begin{table}[!h]
\centering
\small
\begin{tabular}{lcccc}
\toprule
Model                  & FOMC     & NIFTY    & MMLU-F   & AVG      \\
\midrule
Llama-3.1-70B          & 50.96    & 31.64    & 62.21    & 48.27    \\
+ \TELLME              & 52.88    & 31.37    & 63.81    & 49.35    \\
\bottomrule
\end{tabular}
\caption{Performance of \TELLMES on a Large-Scale Model}
\label{tab:appendix_model_size}
\end{table}
Table~\ref{tab:appendix_model_size} reports the performance of \TELLME\ on a 70B-parameter model trained with 4-bit quantization and Low‑Rank Adaptation (LoRA, rank=16) to fit within our computational budget. The method still achieves an improvement of about 1.08 points, indicating that \TELLME\ remains effective even at a much larger scale.

\subsection{Cross-Lingual Transfer and Korean Adaptation}
\label{app:tellme-ko}



\paragraph{Cross-lingual Transfer via \TELLMES Framework.}
We further examined whether the proposed \TELLMES framework generalizes beyond English. To this end, we selected Korean, a language that is linguistically and typologically distinct from English in both grammar and character system, to evaluate its cross-lingual scalability. Due to licensing constraints on Korean financial corpora, we could not directly use domain-specific financial datasets. Instead, we generated Korean \TELLMES data by translating and adapting the English seed corpus(Bloomberg).

Specifically, we created two bilingual variants to explore different degrees of linguistic transfer:
\begin{itemize}
    \item \textbf{\textsc{tellme-bi}}: which preserves English passages but provides Korean question–answer pairs, and
    \item \textbf{\textsc{tellme-ko}}: a fully translated version where both passages and QA pairs are in Korean.
\end{itemize}
The detailed prompt design and example outputs used for this data construction are provided in Table \ref{tab:colab-prompt-simple} and Table \ref{tab:colab-example}, respectively.

\paragraph{Experimental Setup and Evaluation.}
Following the same procedure used for the English data construction, we generated a total of 100,000 samples. The evaluation was conducted using the KoBEST~\cite{jang2022kobest} benchmark, which enables a comprehensive assessment of both Korean linguistic competence and reasoning ability across multiple subtasks. We evaluated Korean performance based on the OLMo2-1B and OLMo2-7B models, aiming to measure how effectively the proposed \TELLMES framework can enhance Korean proficiency in models that originally lack any Korean capability.

\paragraph{Overall Improvement on KoBEST Benchmarks.}
Table 15 summarizes the results on the KoBEST benchmark suite, covering BoolQ, COPA, Hellaswag (HLSW.), Sentineg (SENT.), and WiC. Across all tasks, \textsc{tellme-ko} consistently enhances the base model’s accuracy. For the smaller OLMo2-1B model, \textsc{tellme-ko} improves the average accuracy from 0.477 (Base) to 0.522 in the 0-shot setting, a relative gain of +4.5 \%, and from 0.495 → 0.579 (+8.4 \%) in the 5-shot setting. The larger OLMo2-7B model shows a similar trend, achieving 0.508 → 0.568 (+5.9 \%) in 0-shot and 0.543 → 0.598 (+5.5 \%) in 5-shot evaluation, demonstrating that \TELLMES effectively scales across model sizes.

\begin{table}[!h]
\centering
\scriptsize
\setlength{\tabcolsep}{3pt}
\label{tab:expr_colab_internal}
\begin{tabular}{@{}clcccccc@{}}
\toprule
$\mathbf{N}$ & Setting & BoolQ & COPA & HLSW. & SENT. & WIC & Avg. \\
\midrule
\multicolumn{8}{@{}l}{OLMo2-1B}\\
\midrule
\multirow{2}{*}{0} & Base    & 0.502 & 0.492 & 0.418 & 0.486 & 0.488 & 0.477 \\
\multirow{2}{*} & \textsc{tellme-ko} & 0.632 & 0.534 & 0.418 & 0.511 & 0.517 & \textbf{0.522} \\
\hdashline
\addlinespace[1pt]
\multirow{4}{*}{5} & Base         & 0.522 & 0.477 & 0.422 & 0.524 & 0.529 & 0.495 \\
\multirow{4}{*} & \textsc{cpt}    & 0.571 & 0.632 & 0.534 & 0.536 & 0.515 & 0.546 \\
\multirow{4}{*} & \textsc{tellme-bi} & 0.619 & 0.545 & 0.426 & 0.597 & 0.533 & 0.549 \\
\multirow{4}{*} & \textsc{tellme-ko}   & 0.610 & 0.585 & 0.422 & 0.730 & 0.550 & \textbf{0.579} \\
\midrule
\multicolumn{8}{@{}l}{OLMo2-7B}\\
\midrule
\multirow{2}{*}{0} & Base    & 0.548 & 0.526 & 0.467 & 0.511 & 0.488 & 0.508 \\
\multirow{2}{*} & \textsc{tellme-ko} & 0.625 & 0.535 & 0.490 & 0.650 & 0.541 & \textbf{0.568} \\
\hdashline
\addlinespace[1pt]
\multirow{2}{*}{5} & Base    & 0.726 & 0.548 & 0.440 & 0.511 & 0.488 & 0.543 \\
\multirow{2}{*} & \textsc{tellme-ko} & 0.731 & 0.548 & 0.456 & 0.738 & 0.515 & \textbf{0.598}\\
\bottomrule
\end{tabular}
\caption{Benchmark results on KoBEST for the Olmo-1B and Olmo-7B base models, as well as for models with \textsc{tellme-ko} applied. (N) denotes the number of in-context samples (shots) used for evaluation. The results show the performance of the OLMo2-1B and -7B models, along with the performance when the \textsc{tellme-ko} method is applied. The evaluation metric is accuracy, and the sub-tasks include BoolQ, COPA, Hellaswag (HLSW.), Sentineg (SENT.), and WIC.}
\end{table}

\paragraph{Task-wise Analysis.}
Performance gains vary across task categories. For BoolQ (yes/no comprehension), \textsc{tellme-bi} and \textsc{tellme-ko} exhibit the largest improvement, reaching 0.632 and 0.610 (vs. base 0.502) in the OLMo2-1B 0-shot setting—an absolute increase of over +0.10. This suggests strong transferability in sentence-level reasoning. For COPA, a causal reasoning task, accuracy improves from 0.492 → 0.534 (\textsc{tellme-bi}) and 0.585 (\textsc{tellme-ko}), highlighting enhanced inferential ability after bilingual exposure. In contrast, HellaSwag (commonsense completion) shows minor or negligible gains, implying that narrative completion may require richer Korean pretraining. Notably, Sentineg—a sentiment polarity classification task—benefits substantially from \textsc{tellme-ko}, rising from 0.486 → 0.511 (0-shot) and up to 0.730 (5-shot), showing that cross-lingual alignment improves affective understanding in Korean.
Finally, WiC, which tests semantic consistency of word senses, exhibits moderate but stable improvements (+0.02–0.04 absolute).

\paragraph{Effect of Bilingual vs. Fully Translated Data.}
Comparing \textsc{tellme-bi} and \textsc{tellme-ko} provides insight into the nature of cross-lingual transfer. The bilingual setup (English passages with Korean QA) yields strong improvements in sentence understanding (BoolQ, COPA), indicating that exposure to mixed-language contexts suffices for semantic alignment. However, \textsc{tellme-ko}, which offers fully localized Korean data, surpasses \textsc{tellme-bi} in most settings, especially in the 5-shot Sentineg and COPA tasks, demonstrating that full translation amplifies Korean adaptation while maintaining English-aligned reasoning ability. This suggests that bilingual and translated data jointly facilitate smoother cross-lingual transfer.

\paragraph{Observation on LLama-3.2 in Korean.}
As shown in Table~\ref{tab:tellme_llama_kobest}, applying the \textsc{tellme-ko} framework to LLaMA-3.2-1B results in modest yet consistent improvements in Korean performance on KoBEST. The overall average increases from 0.474 to 0.481, with notable gains in BoolQ (+0.027) and Sentineg (+0.030), which test sentence comprehension and sentiment reasoning, respectively. These improvements indicate that cross-lingual exposure through TELLME allows the model to internalize Korean sentence-level semantics without explicit Korean pretraining.

\begin{table}[!h]
\centering
\scriptsize
\setlength{\tabcolsep}{3pt}
\begin{tabular}{lcccccc}
\toprule
\textbf{Model} & \textbf{BoolQ} & \textbf{COPA} & \textbf{HLSW.} & \textbf{SENT.} & \textbf{WiC} & \textbf{Avg.} \\
\midrule
\textsc{LLaMA3.2-1B} & 0.499 & 0.525 & 0.390 & 0.469 & 0.488 & 0.474 \\
\textsc{tellme-ko} & \textbf{0.526} & 0.520 & 0.372 & \textbf{0.499} & 0.488 & \textbf{0.481} \\
\midrule
\textbf{Δ (Change)} & +0.027 & -0.005 & -0.018 & +0.030 & +0.000 & +0.007 \\
\bottomrule
\end{tabular}
\caption{Comparison of KoBEST performance between \textsc{LLaMA-3.2-1B} (Original) and \textsc{tellme-ko}. The evaluation metric is accuracy. \textsc{tellme-ko} shows consistent gains in BoolQ and Sentineg, indicating improved comprehension and sentiment reasoning ability in Korean.}
\label{tab:tellme_llama_kobest}
\end{table}

\begin{table}[h]
\centering
\scriptsize
\begin{tabularx}{\columnwidth}{>{\raggedright\arraybackslash}X}
\toprule
\textbf{System message} \\
\midrule
Task: Read the text, ask \textbf{Korean} questions and answer them. \newline
Follow these instructions: \newline
1. Ask diverse questions\newline
2. Focus on asking questions about factual information, important knowledge\newline
3. Write questions and answers in \textbf{Korean}\newline
4. Tag the question with ``질문:" and the answer with ``답변:". \newline

Task: \newline
After reading the above text, ask up to 8 questions and provide the correct answers. \newline

Format:\newline
- 질문: [first question] 답변: [first answer]\\
\bottomrule
\end{tabularx}
\caption{Prompt for generating \textsc{tellme-bi} data using English seed passages.
This dataset connects English knowledge with Korean question–answer pairs, enabling the model to jointly learn information across both languages.}
\label{tab:colab-prompt-simple}
\end{table}

\begin{table}[!h]
\scriptsize
\begin{tabularx}{\columnwidth}{>{\raggedright\arraybackslash}X}
\toprule
\textbf{Plain Text} \\
\midrule
Sure, data modeling is the first step to designing a database, but that’s just a coincidence. A well designed database is well designed both because it’s efficient and because it matches business needs.\\
The first step in data modeling is understanding entities. An entity is like a business object: examples may include customer, order, product, patient, blogger, post, or whatever.(...)\\
\midrule
\textbf{Korean QA} \\
\midrule
- \textbf{질문}: 데이터 모델링이 반드시 데이터베이스 설계에만 관련 있나요? \textbf{답변}: 아니요, 데이터 모델링은 비즈니스 프로세스를 이해하고 개선하는 데도 중요한 역할을 합니다.\\
- \textbf{질문}: 엔티티는 무엇을 의미하나요? \textbf{답변}: 엔티티는 고객, 주문, 제품 등과 같은 비즈니스 객체를 의미합니다.\\
- \textbf{질문}: (...) \\
\bottomrule
\end{tabularx}
\caption{Example of data generated based on the Korean version of \textsc{tellme-bi}. Each sample consists of an English passage and a Korean Q\&A pair. An additional version, \textsc{tellme-ko}, was produced by translating the English passages into Korean so that the entire sample (passage–question–answer) is in Korean.}
\label{tab:colab-example}
\end{table}

\end{document}